%% file: main.tex
\documentclass[numsec,webpdf,modern,medium,namedate]{oup-authoring-template}

\onecolumn

\usepackage{booktabs}
\usepackage{tabularx}
\usepackage{makecell}
\usepackage{cprotect}

\graphicspath{{figs/}}

\theoremstyle{thmstyleone}%
\newtheorem{theorem}{Theorem}
\theoremstyle{thmstyletwo}%
\theoremstyle{thmstylethree}%

\usepackage[doublespacing]{setspace}
\usepackage[fontsize=12pt]{fontsize}

\begin{document}

\journaltitle{Journals of the Royal Statistical Society}
\DOI{DOI HERE}
\copyrightyear{XXXX}
\pubyear{XXXX}
\access{Advance Access Publication Date: Day Month Year}
\appnotes{Original article}

\firstpage{1}


\title[SurvSurf]{SurvSurf: a partially monotonic neural network for first-hitting time prediction of intermittently observed discrete and continuous sequential events}

\author[1]{Yichen Kelly Chen\ORCID{0009-0001-8412-3141}}
\author[1,2]{Sören Dittmer\ORCID{0000-0003-2919-4956}}
\author[3]{Kinga Bernatowicz}
\author[3]{Josep Arús-Pous\ORCID{0000-0002-9860-2944}}
\author[4]{Kamen Bliznashki}
\author[5]{John Aston\ORCID{0000-0002-3770-9342}}
\author[6]{James H.F. Rudd\ORCID{0000-0003-2243-3117}}
\author[1]{Carola-Bibiane Schönlieb\ORCID{0000-0003-0099-6306}}
\author[6]{James Jones}
\author[1,6,$\ast$]{Michael Roberts\ORCID{0000-0002-3484-5031}}

\authormark{Chen et al.}

\address[1]{\orgdiv{Department of Applied Mathematics and Theoretical Physics}, \orgname{University of Cambridge}, \orgaddress{\state{Cambridge}, \country{UK}}}
\address[2]{\orgdiv{ZeTeM}, \orgname{University of Bremen}, \orgaddress{\state{Bremen}, \country{Germany}}}
\address[3]{\orgdiv{Data Science and AI}, \orgname{Evinova, AstraZeneca}, \orgaddress{\state{Barcelona}, \country{Spain}}}
\address[4]{\orgdiv{Data Science and AI}, \orgname{Evinova, AstraZeneca}, \orgaddress{\state{Gaithersburg, MD}, \country{USA}}}
\address[5]{\orgdiv{Department of Pure Mathematics and Mathematical Statistics}, \orgname{University of Cambridge}, \orgaddress{\state{Cambridge}, \country{UK}}}
\address[6]{\orgdiv{Department of Medicine}, \orgname{University of Cambridge}, \orgaddress{\state{Cambridge}, \country{UK}}}

\corresp[$\ast$]{Department of Applied Mathematics and Theoretical Physics, University of Cambridge, CB3 0WA, Cambridge, UK. \href{ mr808@cam.ac.uk}{mr808@cam.ac.uk}}

\received{Date}{0}{Year}
\revised{Date}{0}{Year}
\accepted{Date}{0}{Year}



\abstract{We propose a neural-network based survival model (SurvSurf) specifically designed for direct and simultaneous probabilistic prediction of the first hitting time of sequential events from baseline. Unlike existing models, SurvSurf is theoretically guaranteed to never violate the monotonic relationship between the cumulative incidence functions of sequential events, while allowing nonlinear influence from predictors. It also incorporates implicit truths for unobserved intermediate events in model fitting, and supports both discrete and continuous time and events. We also identified a variant of the Integrated Brier Score (IBS) that showed robust correlation with the mean squared error (MSE) between the true and predicted probabilities by accounting for implied truths about the missing intermediate events. We demonstrated the superiority of SurvSurf compared to modern and traditional predictive survival models in two simulated datasets and two real-world datasets, using MSE, the more robust IBS and by measuring the extent of monotonicity violation.}


\maketitle

\section{Introduction}
\input{text/intro}

\section{Datasets}\label{secData}
\input{text/data}

\section{Method}\label{secMethod}
\input{text/method}

\section{Results}\label{secResults}
\input{text/results}

\section{Discussion}
\input{text/discussion}

\section{Conclusion}
\input{text/conclusion}


\begin{appendices}
\input{text/proof__layer_spec_monotone}

\end{appendices}

\section{Competing interests}
No competing interest is declared.

\section{Author contributions statement}
Y.K.C., S.D. and M.R. conceived the research question. Y.K.C led the overall study design, developed the methodology, implemented the experiments, performed data analysis, and drafted the manuscript. S.D. and M.R. supported the theoretical derivations.  S.D., K.B., J.A.P., K.B., J.J. and M.R. provided critical feedback on the experimental design and the interpretation of results.  All authors discussed the findings and approved the final version of the paper. 

\section{Acknowledgments}
The authors are grateful for the following funding: AstraZeneca (YKC), the EU/EFPIA Innovative Medicines Initiative project DRAGON (101005122) (C-BS, MR), the Trinity Challenge (C-BS, MR), the EPSRC Cambridge Mathematics of Information in Healthcare Hub EP/T017961/1 (JA, C-BS, JR, MR), the Cantab Capital Institute for the Mathematics of Information (C-BS), the European Research Council under the European Union’s Horizon 2020 research and innovation programme grant agreement no. 777826 (C-BS), the Alan Turing Institute (C-BS), Wellcome Trust (JR), Cancer Research UK Cambridge Centre (C9685/A25177) (C-BS), British Heart Foundation (JR), the NIHR Cambridge Biomedical Research Centre (JR), HEFCE (JR). In addition, C-BS acknowledges support from the Leverhulme Trust project on ‘Breaking the non-convexity barrier’, the Philip Leverhulme Prize, the EPSRC grants EP/S026045/1 and EP/T003553/1 and the Wellcome Innovator Award RG98755.

\bibliographystyle{abbrvnat}
\bibliography{reference}

\clearpage
\section{Supplemental Material}
\input{text/supp}

\end{document}

%% file: text/intro.tex
Survival models that characterize a single event have seen a wide range of applications in healthcare \citep{RN6}, engineering \citep{RN1,RN5} and social sciences \citep{RN4,RN3}. These models estimate the effect of predictor variables on the distribution of event time to predict the probability of an event (e.g. death) occurring before a certain time (i.e. 1 - the survival probability $S(t)$), where time is measured from a common baseline e.g. since birth or since the initiation of a treatment. Many scenarios involve multiple interdependent events of interest. While certain types of interdependency, such as competing risks \citep{RN7, RN25, RN8} have been studied extensively, the interdependence between sequential events remains under-explored.

Events are considered sequential when the first occurrence (i.e. first hitting time) of the $k$ th event has to precede the occurrence of the $k+1$ th event. For example, a crack cannot reach $k$ mm long without having reached $k-1$ mm, which can only occur after it has grown to $k-2$ mm in length. Similarly, for medical conditions that can progress through different severity grades, the first hitting time for grade $k+1$ can only occur after the first hitting time for grade $k$, even if the less severe state may not be explicitly observed and the person may fluctuate between severity grades. Another less obvious but directly relevant scenario is recurrent events, where the time to the $k+1$ th recurrence must be later than the time to the $k$ th recurrence. 

Due to such interdependence, sequential events should not be modeled independently from each other, as the distribution of the first hitting time ($1 - S(t) = CIF(t)$, cumulative incidence function without competing events) of an earlier-order event would inform that of a later-order event and vice versa. More specifically, $CIF$s of sequential events have a monotonic relationship, where $CIF(t)$ of the $k+1$ th event must be lower than that of the $k$ th event because of its later occurrence. This should not be violated in model prediction, as a later event cannot occur until an earlier event has occurred.

There has been no study aiming at predicting the $CIF$ of sequential events from a set of baseline predictors, but several studies have investigated other related aspects of sequential events. Linear models evolved from Cox regression have been proposed for estimating the effect of independent variables on the hazard function of sequential (recurrent) events \citep{RN22, RN21, RN23}; There are methods for estimating the $CIF$s in a homogeneous population without considering the influence from any independent variables \citep{RN28, RN10}. 
There are also multi-state models that attempt to estimate how events evolve from one to another \citep{RN24, RN19}.  By collecting enough observations and by correctly assuming that the trajectories follow certain random processes, it may be possible to fit an accurate model that describes the transitioning dynamics, and then integrate from this dynamic the marginal distribution of first hitting times. However, this process is not trivial, especially when both time and events can be discrete or continuous, and when unknown nonlinear relationship between features, event order or grade ($g$) and time ($t$) and transitioning dynamics is expected. The multi-state models are also unable to handle unobserved intermediate events. Hence, we have limited the scope of this work to directly predicting the marginal distribution (i.e. without conditioning on past events), rather than through modeling of event transitions.

Importantly, although theoretical guarantees for the monotonic relationship between the predicted $CIF$s may be achieved by including event order as a covariate in linear Cox models, in nonlinear models e.g. DeepHit \citep{RN25} or random survival forests \citep{RN11} there is no such guarantee. 

Another challenge in predicting the $CIF$s of sequential events is when there is a lack of observation or recording on intermediate events due to e.g. rapid progression, so only the more severe form of a disease is recorded. When an event in the sequence is not observed, this does not mean a complete lack of information for the event. Observations on later and earlier events in the sequence imply whether the missing event has or has not occurred by the observation times. This ‘implicit truth’ on missing intermediate events is not accounted for in existing model evaluation metrics such as integrated Brier score (IBS) \citep{RN13} or time-dependent receiver operating characteristic analysis \citep{RN14}, nor in existing modeling approaches.

We therefore propose a neural-network based survival model (SurvSurf) specifically designed to tackle the above-mentioned challenges and gaps in simultaneous time-to-event modeling of sequential events. We demonstrate a theoretical guarantee for the monotonic relationship between the $CIF$s of sequential events while nonlinear influence from predictors is still allowed. SurvSurf also incorporates the implicit truth for unobserved intermediate events in model fitting and unifies both discrete and continuous threshold-defined events. The compatibility with both discrete and continuous time, as well as with discrete and continuous (i.e. threshold-defined) events was made possible through query-able model outputs. The output evaluated at $t$ and $g$ represents the probability of event $g$ occurring by time $t$, and $g$ is a non-negative number whose value is proportional to the order of the event of interest. We use $g=0$ to represent when none of the events have occurred. Depending on the application, event $g$ could mean, e.g. `progressing to a disease severity grade $g$', `(first-time) reaching a maximum price $g$', `having $g$ recurrences' etc.

We also propose a modified IBS, named implied-truth-imputed integrated Brier score ($IBS^{ipcw}_{iti}$), which incorporates the implied truth about the missing intermediate events in computing the IBS. The proposed score showed a very strong correlation with mean squared error ($MSE$) when comparing the true and predicted $CIF$s on simulated data.

The proposed model was evaluated with simulated (Sim-Main, Sim-LackProg) and real disease progression data (RW-TRAE), to predict the probability of reaching a maximum disease severity grade $g$ by time $t$. All subjects are monitored intermittently. Progression to more severe grades was reported/recorded as early as possible. Monitoring time-points without progression were retained as well. Subjects can be right-censored. The model was also applied to UK property price data from 2015 to 2022 (RW-Property), with the aim of predicting whether unseen property types at different locations will see a maximum price increase of at least $g$\% by $t$ years since 2015. In all datasets, SurvSurf demonstrated superior predictive performance and monotonicity preservation compared to all other methods.

%% file: text/data.tex
\subsection{Simulated data}
Data simulated from Markov chains were used to compare the proposed and existing models on their ability to learn nonlinear relationships between the predictors and $CIFs$. The Markov chains were parameterized to mimic the progression of a disease, where temporary recovery may be observed.

For the main simulation (Sim-Main), observations for 4000 independent subjects were generated. Each subject has a feature vector $x_i\in\mathbb{R}^{32{\times1}}$ that is randomly drawn from independent standard normal distributions. The last feature was binarized so that positive values were replaced with 1 and non-positive values were replaced with 0, to represent biological sex common in medical data. 

\begin{figure*}
  \centering
  \includegraphics[width=1\linewidth]{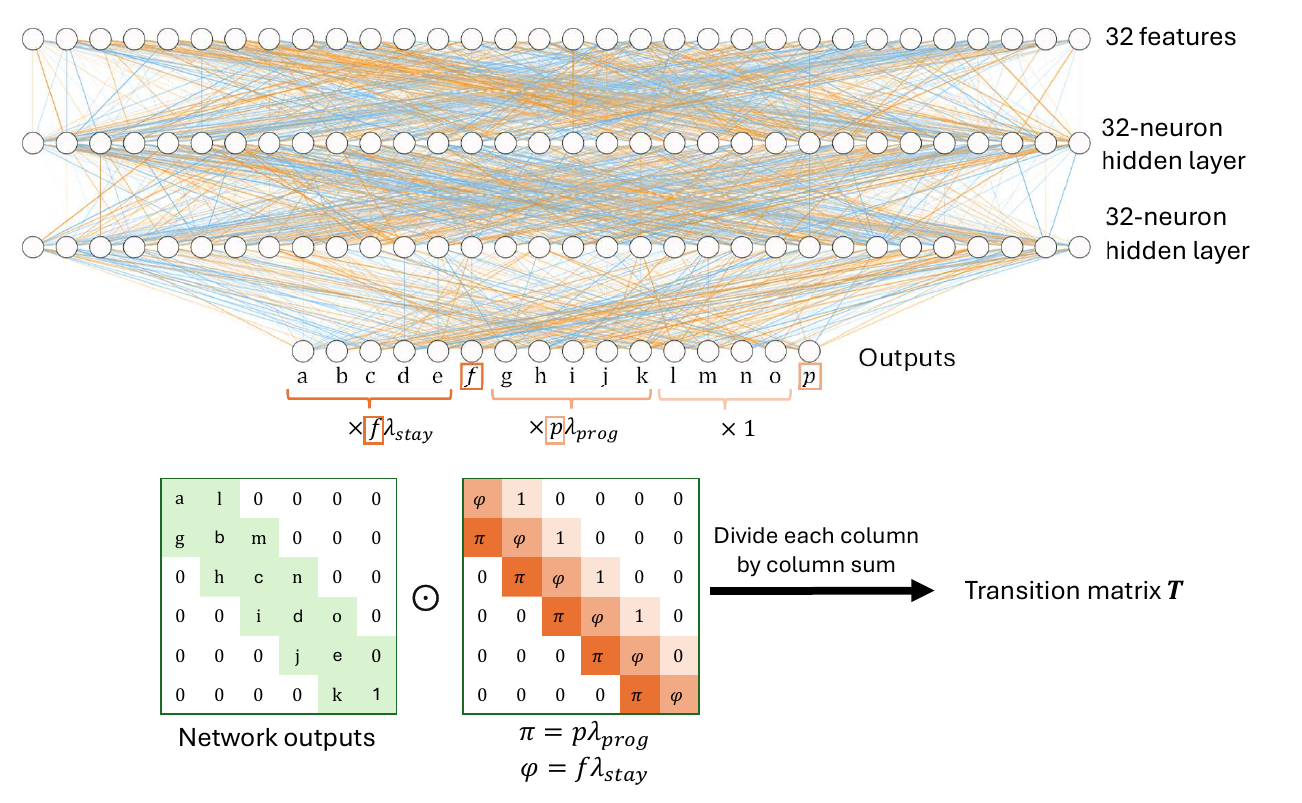}
  \caption{For simulated data, the transition matrix for a subject is generated by passing the feature values of this subject through a neural network. We used $\lambda_{prog} = 2$ and $\lambda_{stay}=1$ for the main simulated dataset (Sim-Main). A smaller dataset with little or no progression (Sim-LackProg) was simulated by setting $\lambda_{prog} = 0.02$ and $\lambda_{stay}=1$ }\label{simulation}
\end{figure*}

To simulate a complex nonlinear relationship between the feature vectors and the $\hat{CIF}$ (the predicted $CIF$) of different events (severity grades), a fully-connected neural network with randomly generated weights and biases (three hidden layers) and sigmoid-like activation functions was used to map each feature vector into elements of a six-state transition matrix (severity grades 0,1,2,3,4,5). The first two layers used $tanh$ for activation, and the last layer used the $Sigmoid$ function. The transition matrix only allows for transition between events that are immediately adjacent to each other in a sequence (i.e. grade 3 can only transit to grade 2 or grade 4), with the last event (grade 5) defined as an absorbing state to represent the irreversible end-stage of a disease (Fig \ref{simulation}).

For Sim-Main, a 10-step trajectory was generated for each subject. Intermittency of observations was simulated by randomly discarding half of the time-points in each trajectory. Additionally, right censoring was simulated by randomly discarding the last 1-3 records of the remaining time-points, keeping at least the first two observations. 17\% of subjects had missing observations for at least one intermediate event. 

A smaller secondary dataset (Sim-LackProg) was also simulated in a similar way to represent a population with little or no progression (Fig \ref{simulation}). Unlike Sim-Main, deliberate censoring was not implemented.

The true $CIF$s for each subject were computed by listing all possible trajectories through the events and the probability of each path occurring, the probability was aggregated for all paths that achieved a specific severity by a certain time to obtain the $CIF(t)$ for that severity.

For each of the simulated datasets, 1000 subjects (and their trajectories) were randomly selected to be the training set. 500 subjects (and their trajectories) were randomly selected to be the validation set. Data for the remaining subjects 
 (2500 for Sim-Main, 500 for Sim-LackProg) were used as the test set which was never exposed to the model fitting and selection process (i.e. early-stopping based on validation-set performance).

\subsubsection{Real-world disease severity data (RW-TRAE)}\label{subsec_RW-TRAE_data}
Treatment-related adverse event (TRAE) data and baseline demographic, clinical and biomarker data was obtained for patients in the comparator arm of the clinical trial study NCT00981058 through Project Data Sphere \citep{RN17}. This is a Phase 3 study that investigates the effect of combining Necitumumab with Gemcitabine-Cisplatin chemotherapy (the experimental arm) relative to Gemcitabine-Cisplatin chemotherapy alone (the comparator arm) in patients with stage IV squamous non-small cell lung cancer.

For each subject, the TRAE data for different organs were pooled into a single trajectory that describes the overall maximum TRAE severity (0,1,2,3,4,5) achieved by each adverse event monitoring/reporting time point. The TRAE severity indicates the level of intervention recommended to a patient experiencing an adverse event. We assume that if there is any TRAE development or progression at all, the subject would experience a stepwise progression from the `no intervention needed' state to requiring increasingly more intensive care (until death), even though the intermediate states may not be recorded due to rapid progression.

\begin{figure}
  \centering
  \includegraphics[width=0.6\linewidth]{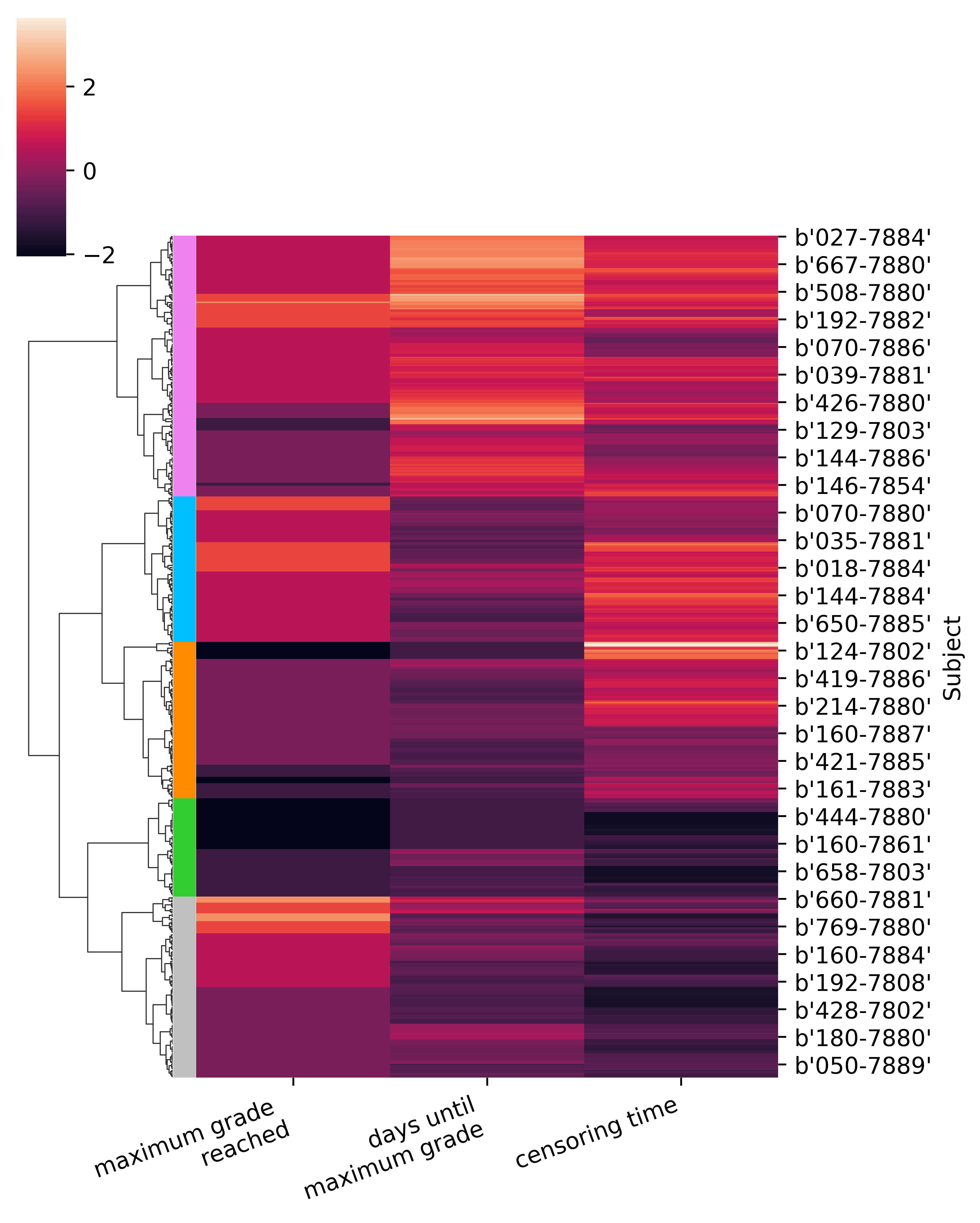}
  \caption{Clustering of subjects in the RW-TRAE dataset derived from the maximum severity grade, time taken to reach the maximum severity, and censoring time. Subject IDs are as provided in the raw data. The values in each column were mean-centered and normalized before plotting. The color gray, blue, orange, pink and green correspond to cluster ID 0, 1, 2, 3 and 4, respectively.}\label{trial_clusters}
\end{figure}

It is not known whether any of the predictors is predictive of the $CIF$s. An artificial predictor was hence injected into the dataset to ensure that there is a known nonlinear association between at least one predictor and the $CIF$s. This enables a model's ability to correctly identify known informative predictors to be assessed. The injected predictor is the cluster ID derived from a hierarchical clustering on the maximum severity reached, the first hitting time for the maximum severity and the time of right-censoring (Fig \ref{trial_clusters}). 

52 baseline variables (including the injected predictor) were included as candidate predictors. 521/549 subjects were retained after removing those who have missing values in more than 5\% of the variables. 56\% of the retained subjects had missing observation for at least one intermediate event. After one-hot encoding of all categorical variables (except for the injected predictor), the remaining missing values were imputed by \verb+KNNImputer+ from \verb+sklearn+.

The dataset was randomly sampled to create the training (321 subjects), validation (100 subjects), and test sets (100 subjects). The sampling was weighted to encourage the training, validation and test sets to all have the same marginal distribution for each baseline variable (See code repository \citet{RN30}).

Non-categorical variables were normalized by applying shifting (subtracting the mean) and scaling (dividing by the standard deviation) parameters derived from the training set. The injected predictor (trajectory cluster ID) was rescaled by dividing by 4.

\subsection{UK property price data (RW-Property)}
Quarterly median property price \citep{RN15} in England and Wales between 2015 and 2022 were combined (by property type and local authority) through a weighted mean into approximate yearly average price (weights are proportional to the number of sales in each quarter). The yearly average was only computed for years with at least 50 sales in total to ensure its representativeness. Such yearly average was available for each build type (new or existing), and property type (terraced, semi-detached, detached and flat/apartment) in 322 local authorities. A trajectory was defined as the percentage change in the yearly average price relative to that of 2015 for a specific property type (accounting for new/existing builds) in a specific local authority. There were a total of 2297 trajectories.

Local and property-type specific characteristics from the 2011 UK Census data \citep{RN16} (tables DC4102EW, DC4402EW, KS601UK, LC2201EW, QS119EW and QS611UK) and the geographical location (longitude and latitude) of each local authority were used as predictors. There were 52 predictors in total (See code repository \citet{RN30}).

The dataset was randomly sampled to create the training (1897 trajectories), validation (200 trajectories), and test sets (200 trajectories). The sampling was weighted to encourage the training, validation and test sets to all have the same marginal distribution for each predictor.

Non-categorical variables were normalized by applying shifting (subtracting the mean) and scaling (dividing by the standard deviation) parameters derived from the training set.

%% file: text/method.tex
\subsection{Proposed model architecture}\label{sec:proposed_model_spec}
The model we propose is an artificial neural network that outputs a probability (i.e. the probability of event $g$ occurring by time $t$) which monotonically increases with $t$ and decreases with $g$, and is not monotonic to input features (candidate predictors). The incorporation of monotonicity was inspired by \cite{RN32}, which introduced a network that is monotonic in $t$ for survival modeling of a single event. Our model (SurvSurf) had to use a very different monotonic transformation to guarantee simultaneous monotonicity in $t$ and $g$.

\begin{figure}
  \centering
  \includegraphics[width=0.6\linewidth]{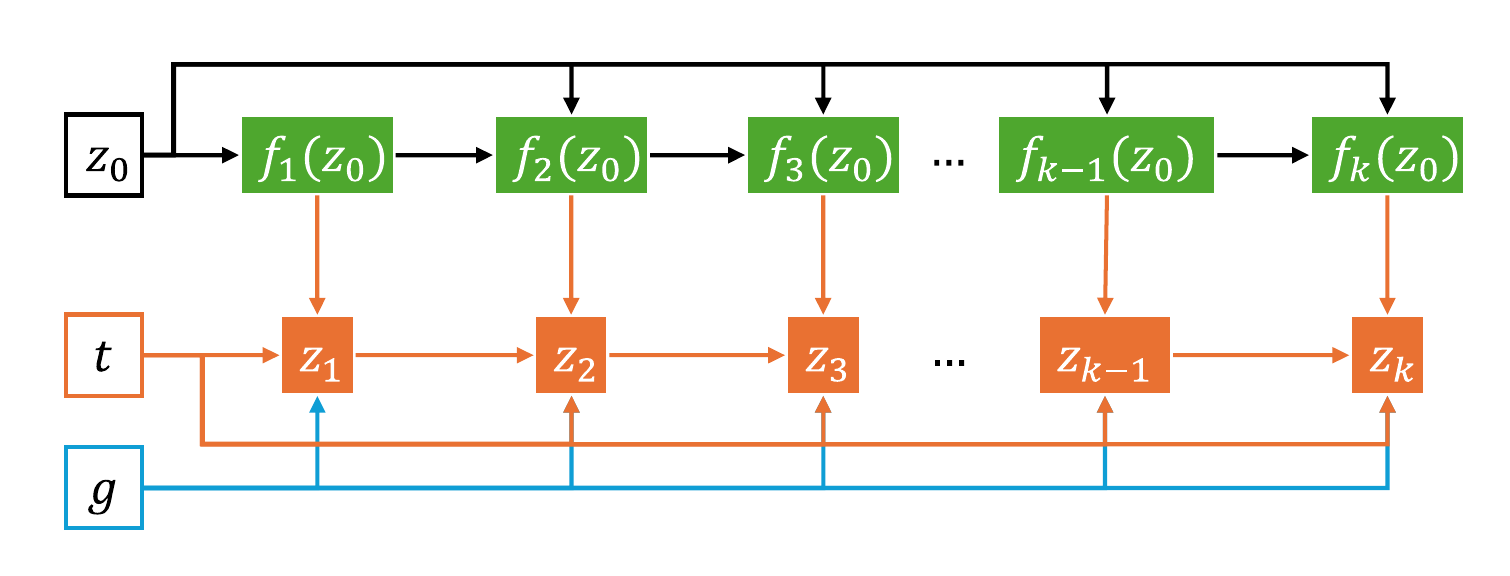}
  \caption{Schematic summary of the SurvSurf network. Orange arrows represent non-decreasing monotonic transformations. Blue arrows represent non-increasing monotonic transformations. Black arrows represent non-monotonic transformations. Symbols are defined in Equation \protect\eqref{eq:z_k_full}.}\label{model_architecture}
\end{figure}

A schematic summary of the network is shown in Fig \ref{model_architecture}. Appendix \ref{secA1} provides a mathematical proof of the monotonicity guarantee of the model output using the chain rule and induction. 

Let $z_{0}$ denote a vector of input features $x$ or a transformation of $x$, the proposed neural network consists of layers ($z_{k}$ for the $k$ th layer) defined as
\begin{equation}\label{eq:z_k_full}
z_{k}(t, g, z_{k-1}, z_0) = \sigma^{layer}_k(\alpha_kt + \gamma_k \diamond h_k(t, g) + A_kz_{k-1} + f_k(z_0) + \beta_k)
\end{equation}
where $\diamond$ is the element-wise multiplication and
\begin{equation}
  \sigma_k^{layer}=
  \begin{cases}
   tanh & \text{if $k<k_{max}$}\nonumber\\
   I & \text{if $k=k_{max}$}\nonumber
  \end{cases}
\end{equation},
\begin{equation}
\begin{split}
h_k(t, g) = [\sigma^{sigm}(M_k^{time}t+c_k^{time})-\sigma^{sigm}(c_k^{time})] \diamond \sigma^{sigm}({-M}_k^{grade}g+c_k^{grade})\nonumber
\end{split}
\end{equation},
\begin{align}
\alpha_k,\gamma_k,M_k^{time},M_k^{grade}&\in\mathbb{R}_{\geq0}^{dim{(z_k)}\times1} \nonumber\\
c_k^{time},c_k^{grade}&\in\mathbb{R}^{dim{(z_k)\times1}} \nonumber\\
A_k&\in\mathbb{R}_{\geq0}^{ dim{(z_k)}\times dim{(z_{k-1})}}\nonumber\\
\beta_k &\in\ \mathbb{R}^{ dim{(z_k)}\times1}\nonumber,
\end{align}
\begin{equation}
  	f_k(z_0) = \sigma^{HardSigm}(C_kf_{k-1}(z_0)+B_kz_0+c_k^{z_0})-0.5\nonumber,
\end{equation}
\begin{align}
  	C_k&\in\mathbb{R}^{dim{(z_k)}\times dim{(z_{k-1})}}\nonumber\\
        f_{0}(z_0)& = 0 \nonumber\\
   	B_k&\in\mathbb{R}^{dim{(z_k)}\times dim{(z_0)}}\nonumber\\
   	c_k^{z_0}&\in\ \mathbb{R}^{dim{(z_k)}\times1}\nonumber,
\end{align}
\begin{align}
  	\text{$\sigma^{sigm}$ is the $Sigmoid$ function} ,
   	\text{$\sigma^{HardSigm}$  is the $Hardsigmoid$ function.} \nonumber
\end{align}
To ensure the output $\hat{CIF}(t=0)=0$ and $0 \leq \hat{CIF} < 1$, the $\hat{CIF}$ is defined as
\begin{equation}
  \hat{CIF}(t,g, z_0) = tanh(z_{k_{max}}(t,g, z_0) - z_{k_{max}}(t=0, g, z_0))
\end{equation}

\subsection{Loss function} \label{sec:LossFn}
We designed a loss function \verb+LossDyDg+ to ensure the network preserves two desirable properties required for sequential events. \verb+LossDyDg+ reflects two facts (truths) implied by the observations of sequential events:
\begin{enumerate}
  \item When an event is observed to be the most severe one before a monitoring time-point $t$ (inclusive), any later event in the sequence (i.e. higher severities, larger $g$) must not have occurred.
  \item If no events have occurred until $t$, then the $\hat{CIF}$ for the earliest event in the sequence (and all later events) must be small.
\end{enumerate}

\begin{equation}
  LossDyDg_{i,t}=-[ylog(\Delta \hat{CIF}_g) + (1-y) log(1-\hat{CIF}(t,g_{i,t},x_i))]
\end{equation}
where
\begin{equation} \label{DyDg_essense}
  \Delta \hat{CIF}_g = \hat{CIF}(t,g_{i,t},x_i) -\hat{CIF}(t,g_{i,t}+\delta_g,x_i)
\end{equation}

For trajectory (or subject) $i$, at each monitoring time-point $t,$ $y_{i,t}$ is a binary indicator for whether the event of interest has occurred. If progression was observed before $t$ (inclusive), then $g_{i,t}=k$ and $y_{i,t}=1$, where $k$ is the most severe grade observed by $t$ (inclusive). Otherwise if there has been no progression from baseline by $t$ (inclusive), i.e. no events occurred by $t$, then $g_{i,t}= 0 + \delta_g$ and $y_{i,t}=0$, where $\delta_g$ is a hyperparameter indicating the minimal difference between events that need to be resolved (if events are defined as a continuous measurement crossing different thresholds). For discrete sequential events where $g_{i,t}$ can only be integers, $\delta_g=1$. 

$LossDyDg_{batch}$ is the mean of $LossDyDg_{i,t}$ for a set of randomly and independently sampled $(i,t)$ that form the batch.
 
\subsection{Model evaluation metrics}
For the simulated data, the true $CIF$s are known, so the the model can be evaluated using $MSE$ between $\hat{CIF}$s and the true $CIF$s at evenly and reasonably spaced time-points for each event. For real-world data, the true $CIF$s are hardly known, so the IBS was used instead.

However, we found that the mean of per-event naive inverse probability of censoring weighted IBS ($IBS^{ipcw}_{naive}$) that completely disregards the implied truths for missing intermediate events has less robust correlation with $MSE$ (Fig \ref{MSE_vs_new_old_IBS_less_prog_sim}). The implied truths are:
\begin{enumerate}
  \item When an event is observed to be the most severe one before a monitoring time-point $t$ (inclusive), any later event in the sequence (i.e. higher severities, larger $g$) must not have occurred (same as Section \ref{sec:LossFn}). 
  \item When a more severe event (larger $g$) is first observed at monitoring time-point $t$, less severe events must have occurred before $t$, even some have not been explicitly observed.
\end{enumerate}

After modifying the $IBS^{ipcw}_{naive}$ to account for the implied truth (named as implied-truth-imputed $IBS^{ipcw}$, $IBS^{ipcw}_{iti}$), a very strong and more robust correlation with $MSE$ was restored (Fig \ref{MSE_vs_new_old_IBS_main_sim} and \ref{MSE_vs_new_old_IBS_less_prog_sim}). The $IBS^{ipcw}_{iti}$ for a single event is defined as 

\begin{equation}\label{IBS_iti_def_lv1}
  IBS^{ipcw}_{iti} = \int^{t_{max}}_{t_{min}}\frac{BS^{ipcw}_{iti}(t)}{t_{max}} dt
\end{equation} and
\begin{equation} \label{IBS_iti_component_BS}
\begin{split}
BS^{ipcw}_{iti}(t) = \frac{1}{N_{certain}(t)} \sum_{i = 1}^{N_{certain}(t)} [\frac{( 0 - \hat{S}(t, x_i))^2 \cdot \mathbb{I}_{T_i \leq t, \varsigma_i = 1}}{ \hat{G}(T_i)}\\ + \frac{ ( 1 - \hat{S}(t, x_i))^2 \cdot \mathbb{I}_{T_i > t}}{ \hat{G}(t)} ]
\end{split}
\end{equation}
where $T_i$ is the first hitting time for the event of interest in subject $i$ or trajectory $i$; $t$ is the time-point to be evaluated; $\hat{G}$ is the Kaplan-Meier estimate for the distribution of whole-trajectory censoring time (i.e. distribution of $C_i$) in the train/validation/test split of interest; $N_{certain}(t)$ is the total number of subjects (or trajectories) for which we are certain whether the event has or has not happened by $t$ based on the observed and the implied truths at $t$. Missing intermediate events will lack these truths for a time interval bounded by the first hitting time of a later-order event and the last observation of an earlier-order event; $\mathbb{I}_{T_i \leq t, \varsigma_i = 1}$ and $\mathbb{I}_{T_i > t}$ are binary indicators encoding the observed and implied truths. 
\begin{equation}
  \hat{S}_{event=k}(t, x_i) = 1 - \hat{CIF}_{event=k}(t, x_i) \nonumber
\end{equation}
\begin{equation}
  \varsigma_i = I(T_i \leq C_i), \text{$C_i$ is the time of the last observation for subject $i$ or trajectory $i$} \nonumber
\end{equation}

Additionally, the extent of monotonicity violation was defined to be the maximum positive value in ${\hat{CIF}}_{event=k+1}(t, x_i)-{\hat{CIF}}_{event=k}(t, x_i)$ across all subjects.

\subsection{Model Benchmarking}
For each dataset, the $IBS^{ipcw}_{iti}$ and the extent of monotonicity violation was compared between SurvSurf,  DeepHit, random survival forest (\verb+RandomSurvivalForest+ in \verb+sksurv+, abbreviated as RSF), gradient-boosted Cox proportional hazard loss with regression trees as base learner (\verb+GradientBoostingSurvivalAnalysis+ in \verb+sksurv+, abbreviated as GBSA), and Cox proportional hazard model (\verb+CoxPHSurvivalAnalysis+ in \verb+sksurv+, abbreviated as CoxPH). Non-Cox models were each trained five times with different random seeds. All existing models were trained by including the event index $g$ (proportional to event order) as an additional covariate (i.e. model input).

DeepHit was translated into \verb|PyTorch|, and was trained with \verb+LossDyDg+ and the SUMO loss. The latter was demonstrated by \cite{RN32} and \cite{RN18} to be a proper scoring rule for survival models. The number of layers, number of neurons and the dropout rate were specified in accordance with the original work. Adam optimizer with a learning rate of 0.0002 and a weight decay of 0.05 was used for model fitting. The weight decay was necessary to achieve convergence in both training and validation loss. The batch size was set to 64 for Sim-Main and Sim-LackProg and RW-Property. For RW-TRAE, the batch size was set to 16. This was to account for the difference in sample size and the potential of overfitting. For the SUMO loss, only the first hitting time for each observed event was used for model training; at the last monitoring time-point $t$, the next event in the sequence which has not occurred yet (maximum $g$ for the trajectory + $\delta_g$) was included for training as a right-censored observation. For \verb|LossDyDg|, all monitoring time-points were used for model training; $\delta_g=1$ for Sim-Main, Sim-LackProg and RW-TRAE; $\delta_g=0.01$ (i.e. 1\% change in property price) for RW-Property.

Similar to DeepHit, the SurvSurf model had 4 layers. For Sim-Main and Sim-LackProg, there were 32 predictors, each SurvSurf hidden layer had 32 units. For RW-TRAE and RW-Property, each layer had 64 units since there were 52 predictors. Adam optimizer with a weight decay of 0.005 was used for model fitting for Sim-Main, Sim-LackProg and RW-TRAE. The weight decay was set to be lower than that for DeepHit due to the presence of monotonicity constraint in SurvSurf. For RW-Property, the weight decay was further reduced to 0.001 as there are many more events (equal to the number of unique values in property price increase) in the training data. The learning rate was set to 0.001 for Sim-Main and Sim-LackProg, and 0.002 for RW-TRAE and RW-Property to balance between potential overfitting and training time. The batch size was set to 64 for Sim-Main, Sim-LackProg and RW-Property. For RW-TRAE, the batch size was set to 16. For \verb|LossDyDg|, all monitoring time-points were used for model training; $\delta_g=1$ for Sim-Main, Sim-LackProg and RW-TRAE; $\delta_g=0.01$ for RW-Property.

All except one hyperparameter for the RSF were set to default. For all datasets, the number of trees was set to 1000.

All except two hyperparameters for the GBSA were set to default. For Sim-Main, Sim-LackProg and RW-Property (i.e. the three larger datasets), the minimum number of samples per leaf was set to 10 and the number of estimators was set to 50. For RW-TRAE, the minimum number of samples per leaf was set to 20 and the number of estimators was set to 50. 

In later sections, a `standard' loss refers to the use of default optimization targets for CoxPH, RSF, and GBSA. CoxPH uses negative partial log-likelihood. The other methods have more complex optimization, details can be found through \verb+sksurv+ \citep{RN31}.

All models that were not based on neural networks used the same training data formatting/transformation as that used for the aforementioned DeepHit-SUMO setup, i.e. only the first hitting time for each observed event was used for model training; and at the last monitoring time-point $t$, the next event in the sequence which has not occurred yet (maximum $g$ for the trajectory + $\delta_g$) was included for training as a right-censored observation.

All DeepHit and SurvSurf experiments were coordinated with \verb|pytorch-lightning| and tracked with \verb|wandb|. Code for the benchmarking experiments is available in code repository \citep{RN30}. Code for the SurvSurf model architecture is available as a separate installable package in code repository \citep{RN29}.

%% file: text/results.tex
\begin{figure}
    \centering
    \includegraphics[width=1\linewidth]{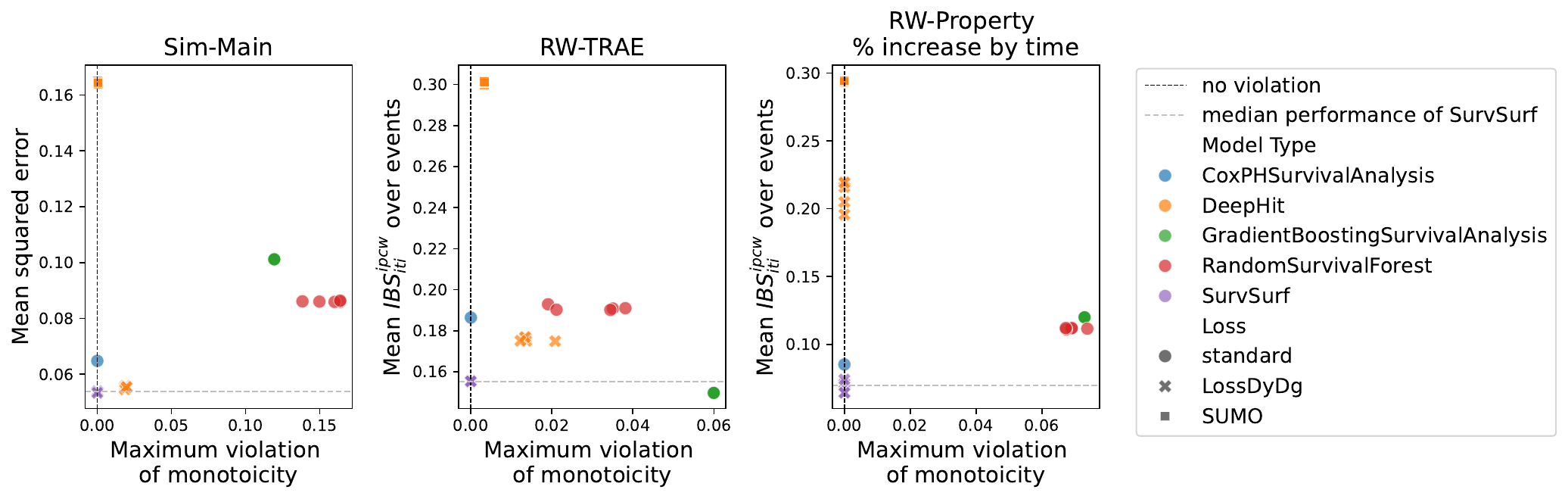}
    \caption{Model performance and the extent of monotonicity violation for the SurvSurf and the benchmarking models in the test set of different datasets. Points with the same color and marker are the same model trained by initializing with different random seeds. Points closer to the bottom-left corner represent models with better performance and less monotonicity violation.}\label{performance_violation_scatter__all_main_ds}
\end{figure}  

\begin{figure}
    \centering
    \includegraphics[width=0.9\linewidth]{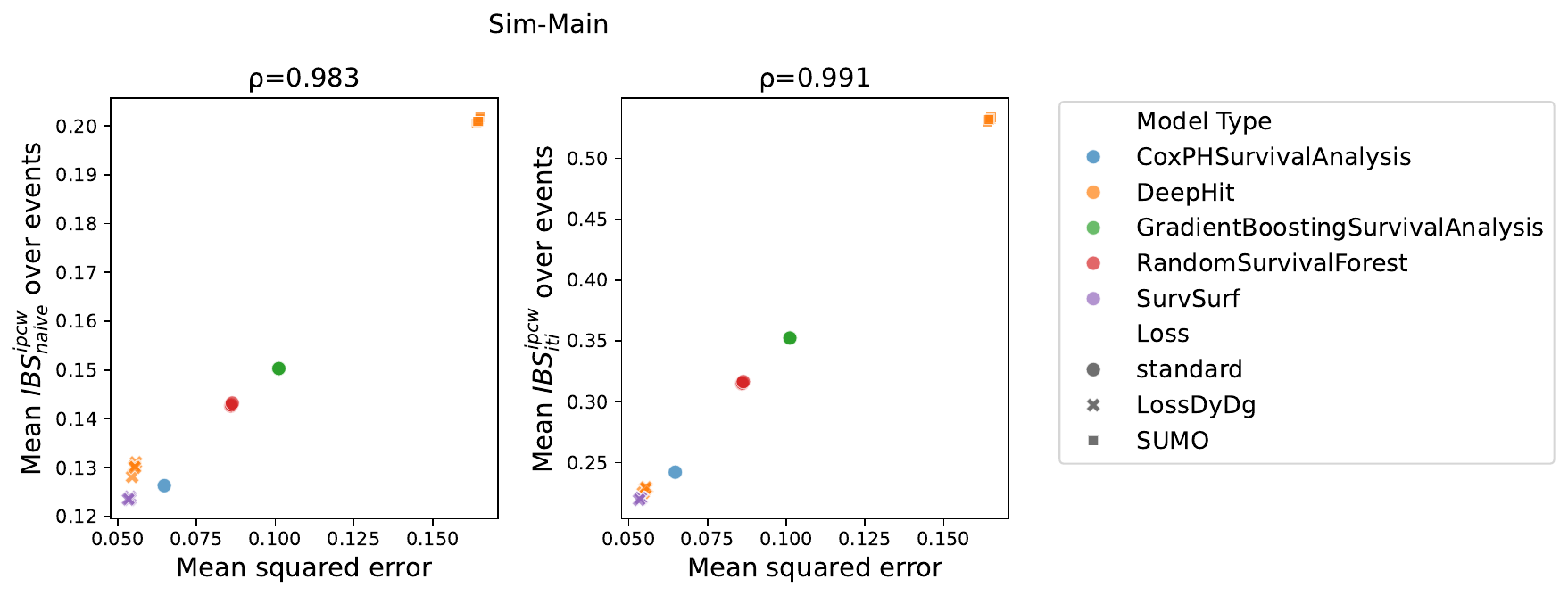}
    \caption{Test-set result: Spearman's rank correlation ($\rho$) between mean squared error and $IBS^{ipcw}_{naive}$ (left) and $IBS^{ipcw}_{iti}$ (right) on Sim-Main dataset.}\label{MSE_vs_new_old_IBS_main_sim}

    \includegraphics[width=0.9\linewidth]{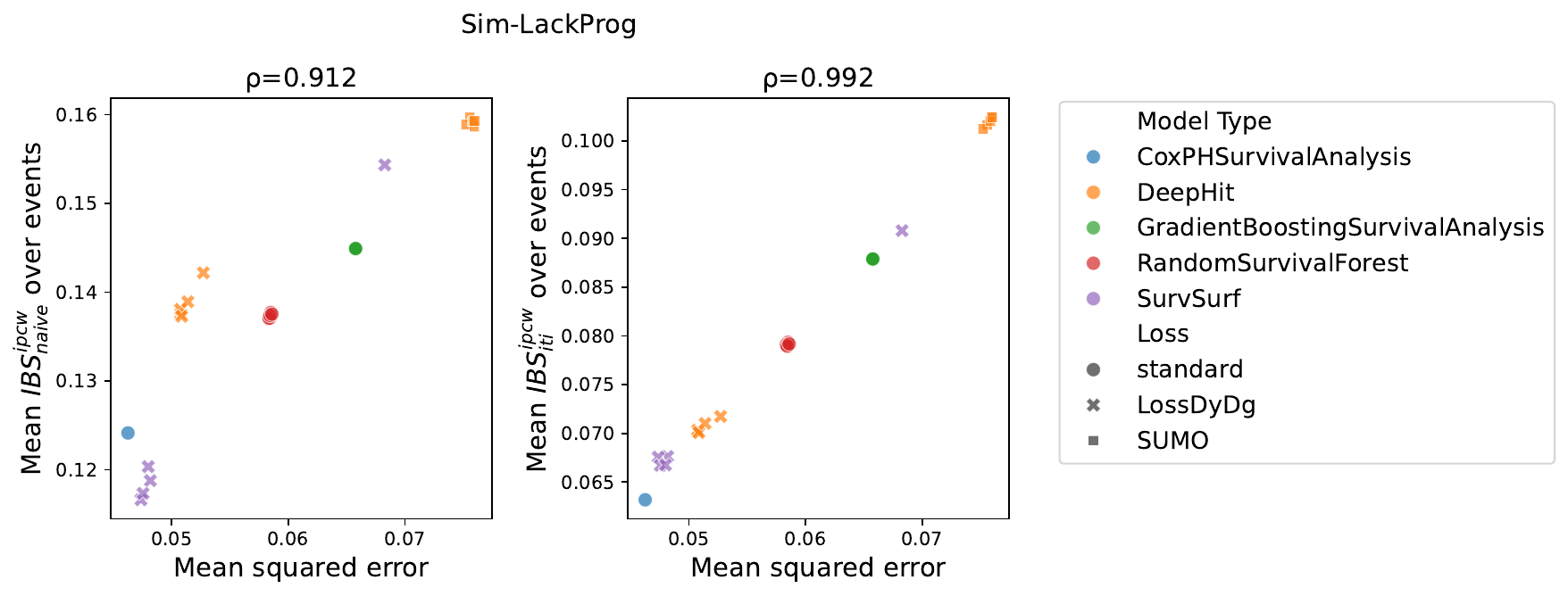}
    \caption{Test-set result: Spearman's rank correlation ($\rho$) between mean squared error and $IBS^{ipcw}_{naive}$ (left) and $IBS^{ipcw}_{iti}$ (right) on Sim-LackProg dataset.}\label{MSE_vs_new_old_IBS_less_prog_sim}
\end{figure}

SurvSurf was compared to modern and traditional survival models in terms of predictive performance and the extent of monotonicity violation. Across datasets, SurvSurf was consistently one of the best-performing models, and was guaranteed to have zero monotonicity violation (also see Appendix \ref{secA1} for proofs of monotonicity).

\subsection{Simulated datasets}
SurvSurf achieved the best performance on Sim-Main (Fig \ref{performance_violation_scatter__all_main_ds} left), i.e. had the smallest mean MSE between the predicted and true $CIF$s computed at grades 1, 2, 3, 4, 5 and across time-points 1-9 (with a spacing of 1). It was also competitive on Sim-LackProg (Fig \ref{MSE_vs_new_old_IBS_less_prog_sim}, Table \ref{table_model_test_set_metrics_simulated}). No monotonicity violation was observed for SurvSurf and all linear models (Table \ref{table_model_test_set_metrics_simulated}). 

The second best model for Sim-Main was DeepHit with \verb|LossDyDg|, which had very similar performance, but was capable of monotonicity violation (Table \ref{table_model_test_set_metrics_simulated}). Other nonlinear benchmarking models had substantially inferior performance and also relatively large monotonicity violation (Fig \ref{performance_violation_scatter__all_main_ds} left, Table \ref{table_model_test_set_metrics_simulated}).  

\verb|CoxPHSurvivalAnalysis| had slightly worse performance than SurvSurf on Sim-Main, but had marginally better performance on Sim-LackProg (Table \ref{table_model_test_set_metrics_simulated}).

\input{tables/table_model_test_set_metrics_simulated}

\subsection{Correlation between $IBS^{ipcw}_{iti}$ and MSE}
In order to assess model performance on real-world data where true $CIF$s are not available, we propose the $IBS^{ipcw}_{iti}$ as a model evaluation metric on real-world data. $IBS^{ipcw}_{iti}$ correlated better with MSE in Sim-Main than $IBS^{ipcw}_{naive}$ did (Fig \ref{MSE_vs_new_old_IBS_main_sim}). In Sim-LackProg,  $IBS^{ipcw}_{iti}$ maintained an almost perfect correlation with MSE, but $IBS^{ipcw}_{naive}$ did not  (Fig \ref{MSE_vs_new_old_IBS_less_prog_sim}). These indicate that $IBS^{ipcw}_{iti}$ is a more robust metric for assessing model prediction. $IBS^{ipcw}_{iti}$ was therefore used for predictive performance evaluation on the two real-world datasets (RW-TRAE and RW-Property). Similar results were also observed without using the inverse-probability-of-censoring weighting (see Supplemental Material \ref{MSE_vs_ipcwless_IBS_iti_main_sim} and \ref{MSE_vs_ipcwless_IBS_iti_less_prog_sim}). In subsequent expriments on real-world data, relative performance between the models were not affected by the presence of absence of the weighting (see Supplemental Material \ref{performance_violation_scatter_ipcwless__all_main_ds} and \ref{performance_violation_scatter_more_g__all_main_ds}).

\begin{figure}
\centering
    \includegraphics[width=1\linewidth]{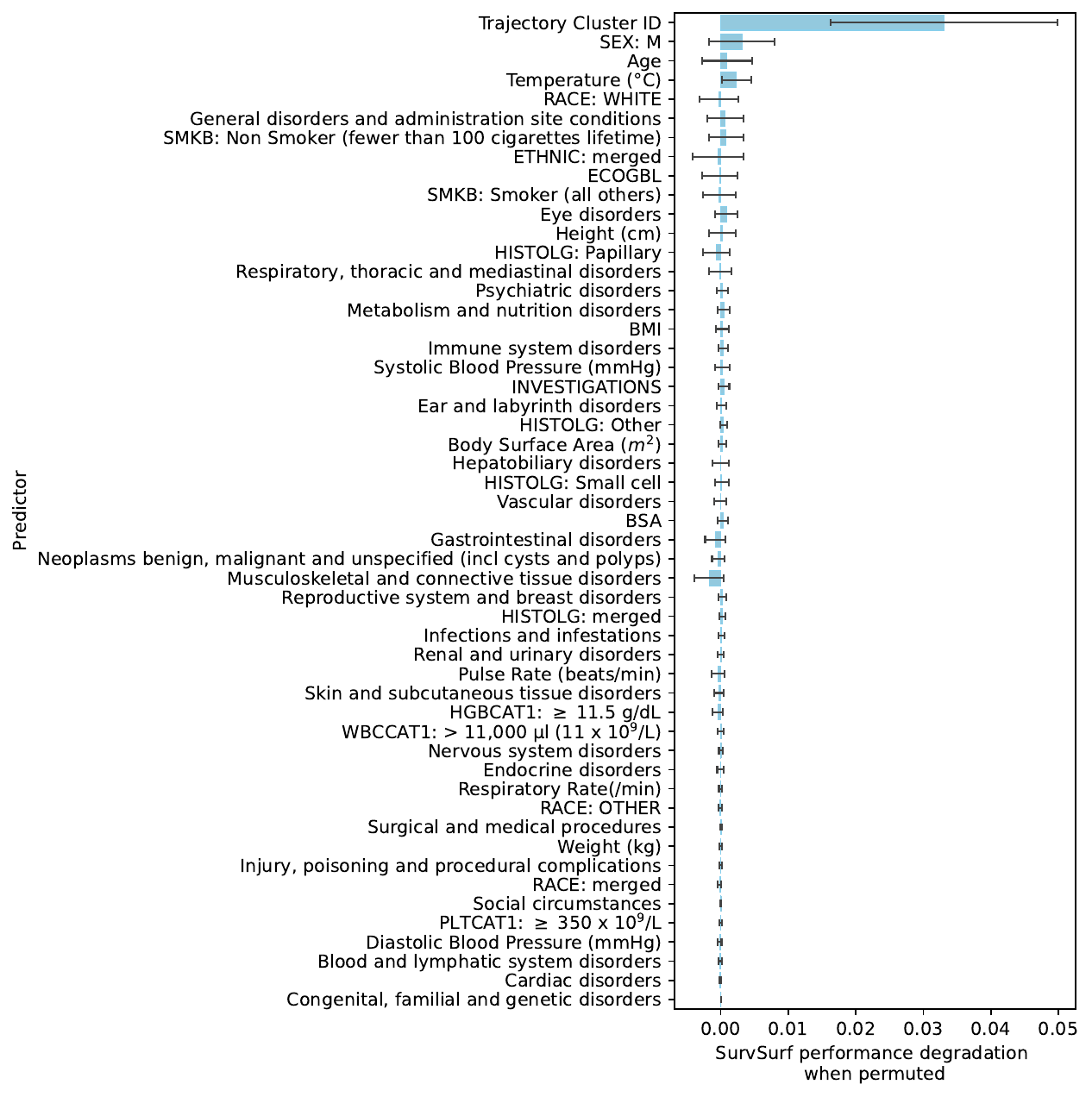}
    \caption{Test-set result: degradation in SurvSurf model performance ($IBI^{ipcw}_{iti}$) caused by independently and individually permuting each predictor in RW-TRAE  50 times. Error bars represent $\pm 1.96 \times SD$.  The larger the performance degradation the more influential a predictor is.}\label{trial_survsurf_feat_perm_perf_change}
\end{figure}

\subsection{Disease progression dataset: RW-TRAE}
In RW-TRAE, approximately 90\% of test-set subjects had stopped being monitored within 140 days from the start of treatment. SurvSurf was competitive to the best-performing model according to the $IBS^{ipcw}_{iti}$ evaluated for $t$s between day 0 and day 140 (with a spacing of 7 days) and averaged across grades 1, 2, 3, 4, 5 (Fig \ref{performance_violation_scatter__all_main_ds} middle). 

SurvSurf had the additional advantage of guaranteed zero monotonicity violation (Fig \ref{performance_violation_scatter__all_main_ds}, Table \ref{table_model_test_set_metrics_real}). The best-performing model was GBSA. Monotonicity violation was observed in all nonlinear models except SurvSurf (Table \ref{table_model_test_set_metrics_real}). 

As expected, the injected predictor (trajectory cluster ID, see Section \ref{subsec_RW-TRAE_data}) was considered to be the most influential predictor by SurvSurf; permuting this predictor resulted in the largest degradation in model performance (Fig \ref{trial_survsurf_feat_perm_perf_change}).

\input{tables/table_model_test_set_metrics_real}

\subsection{Property price dataset: RW-Property}
In RW-Property (Fig \ref{performance_violation_scatter__all_main_ds} right), $IBS^{ipcw}_{iti}$ was evaluated for events defined by property increase 
thresholds ranging from 0\% to 100\% (with 1\% spacing). SurvSurf had the best $IBS^{ipcw}_{iti}$ and had no monotonicity violation (Table \ref{table_model_test_set_metrics_real}). Other non-linear models all had worse performance than CoxPH, which had the second best performance. All tree-based models had noticeable monotonicity violation (Table \ref{table_model_test_set_metrics_real}).

\begin{figure*}
\centering
    \includegraphics[width=1\linewidth]{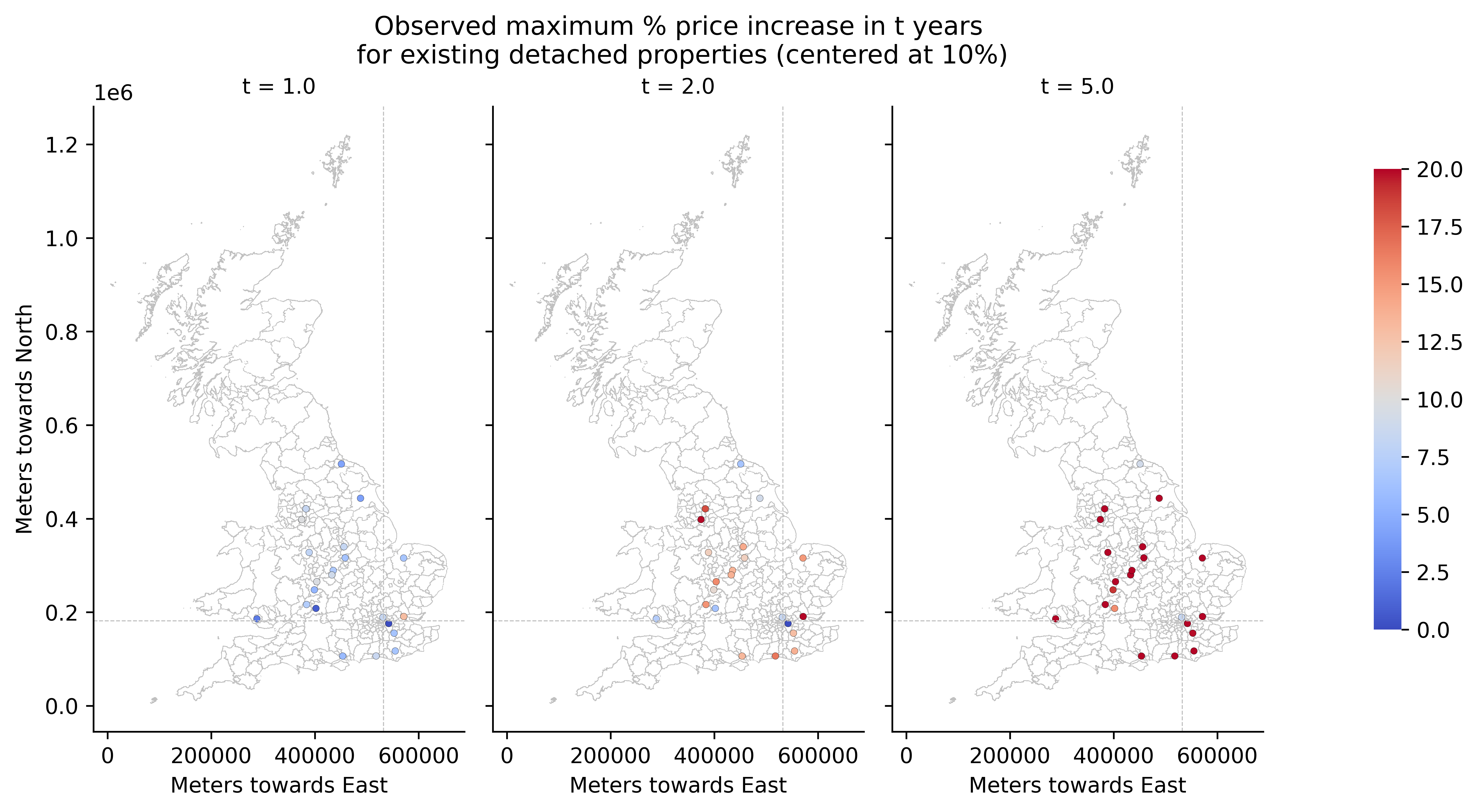}
    \caption{Test-set data: observed maximum percentage increase in the price of existing detached properties at various locations since 2015. The color bar is centered at 10\%, capped at 20\%.}\label{property_price_DetachedExisting_g10_obs}

    \includegraphics[width=1\linewidth]{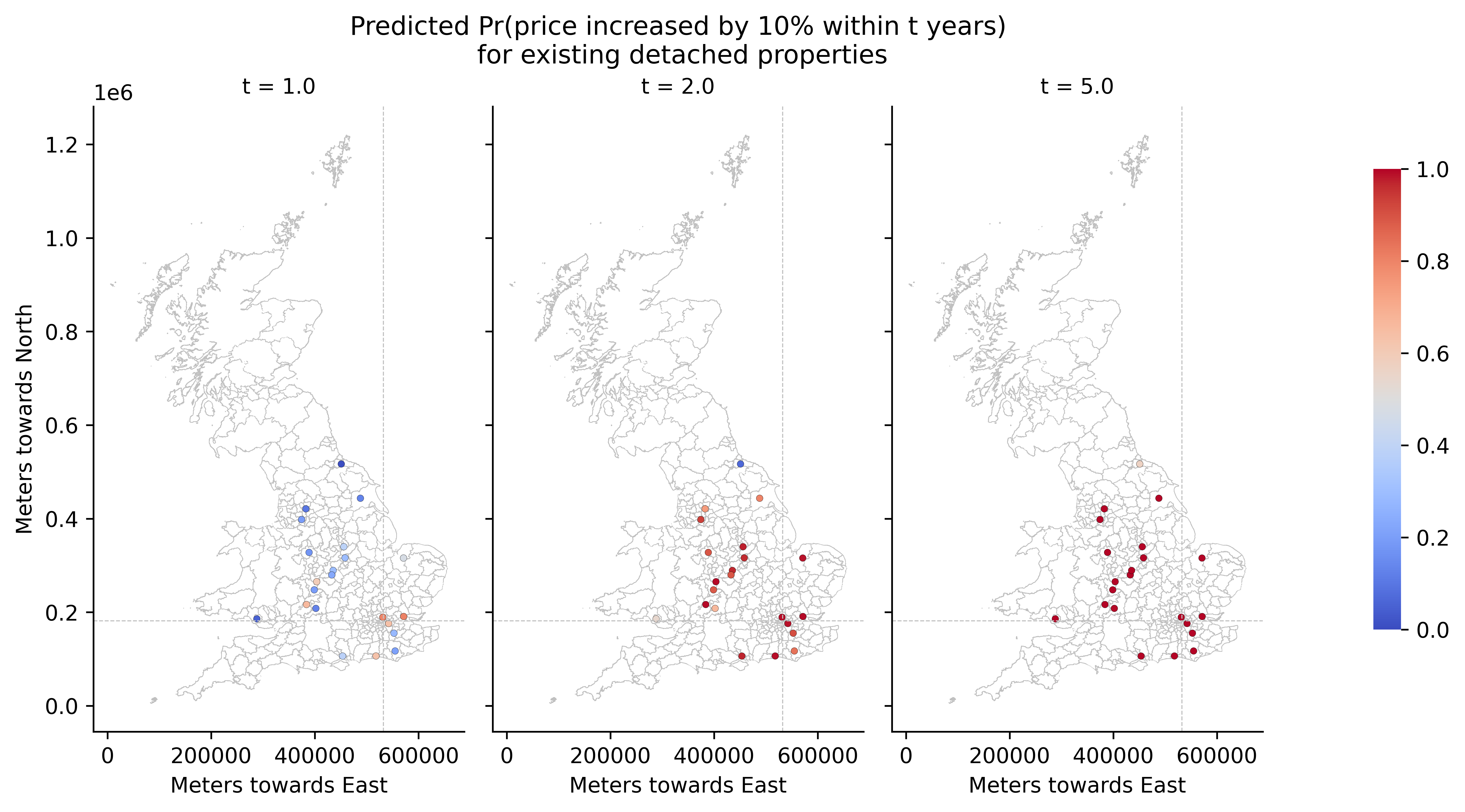}
    \caption{Test-set prediction: predicted probability for reaching a maximum percentage price increase of 10\% by 1, 2, and 5 years since 2015. Results shown are for existing detached properties at various locations. The color bar is centered at 0.5.}\label{property_price_DetachedExisting_g10_pred}
\end{figure*}

\begin{figure*}
\centering
    \includegraphics[width=1\linewidth]{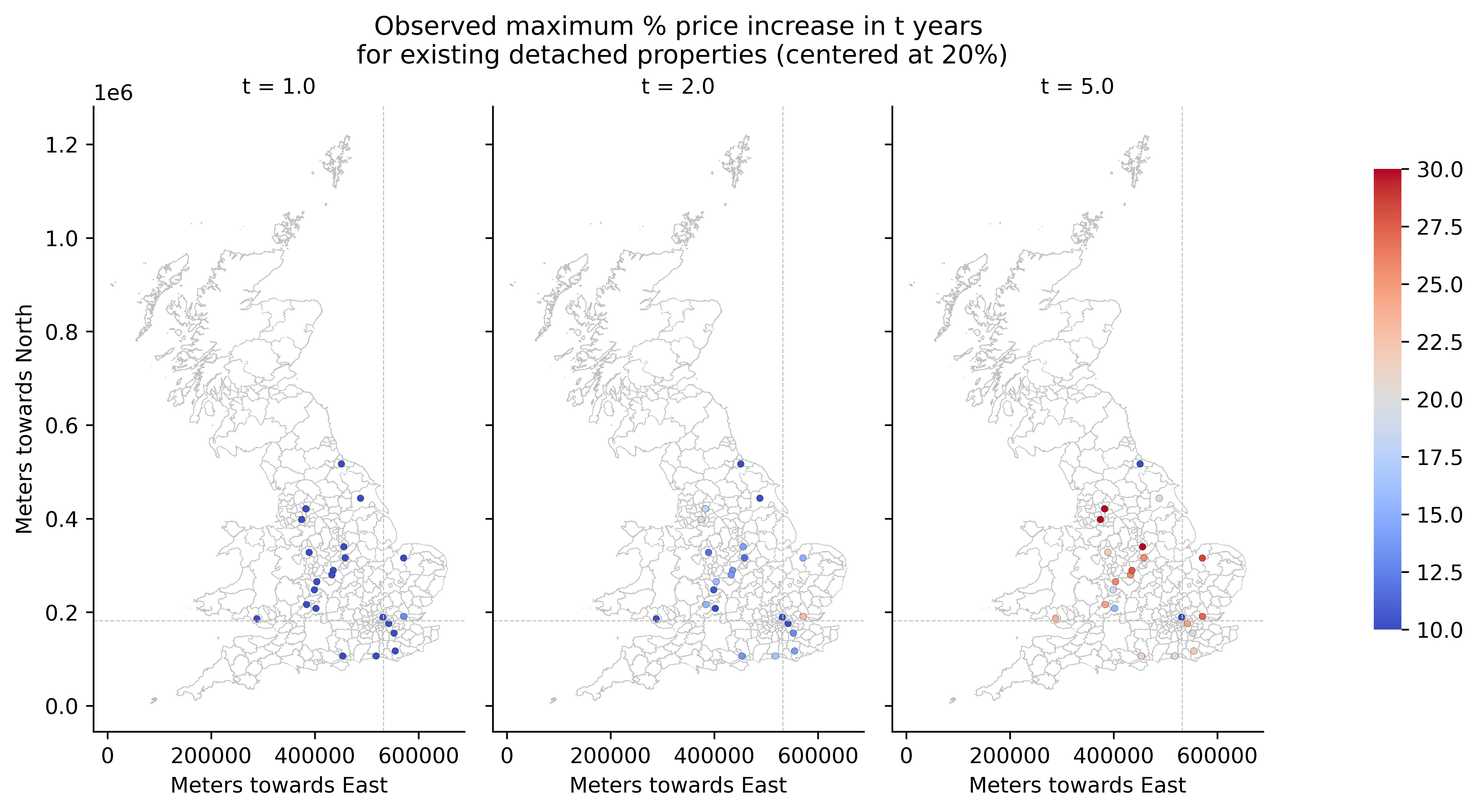}
    \caption{Test-set data: observed maximum percentage increase in the price of existing detached properties at various locations since 2015. The color bar is centered at 20\%, capped at 10\% and 30\%.}\label{property_price_DetachedExisting_g20_obs}

    \includegraphics[width=1\linewidth]{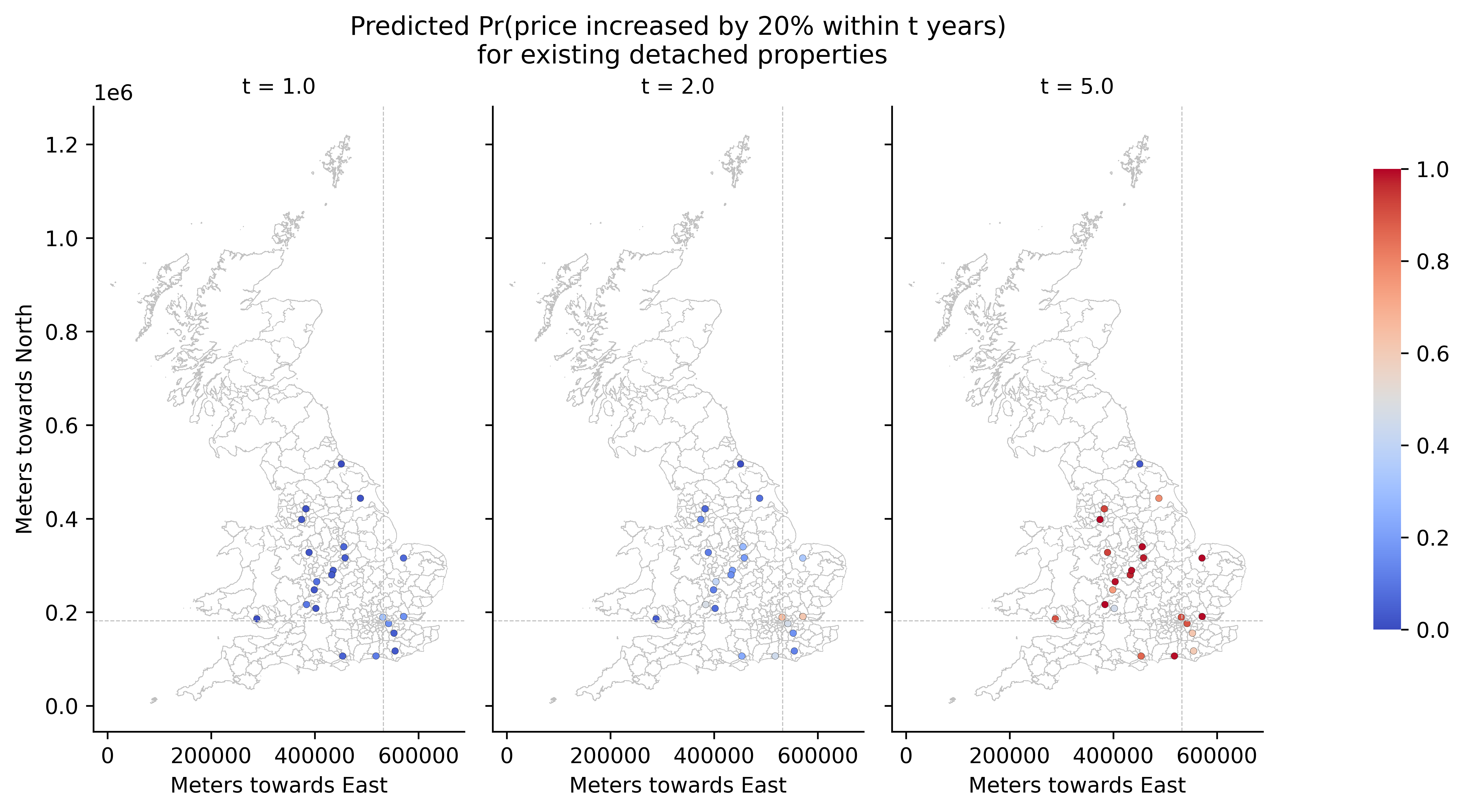}
    \caption{Test-set prediction: predicted probability for reaching a maximum percentage price increase of 20\% by 1, 2, and 5 years since 2015. Results shown are for existing detached properties at various locations. The color bar is centered at 0.5.}\label{property_price_DetachedExisting_g20_pred}
\end{figure*}

An intuitive illustration of SurvSurf prediction on the property price data (test-set) is shown in Fig \ref{property_price_DetachedExisting_g10_obs} to \ref{property_price_DetachedExisting_g20_pred}. The predicted probability of reaching a specific maximum percentage increase in property price by certain time accurately reflects the observed price change. When the maximum price increase was observed to be less than 10\%, the predicted probability of reaching a maximum price increase of 10\% was mostly less than 0.5 (Fig \ref{property_price_DetachedExisting_g10_obs} and  \ref{property_price_DetachedExisting_g10_pred}); and when the maximum price increase was observe to be more than 10\%, the predicted probability was mostly above 0.5. The same pattern was also observed when the price increase threshold was set to 20\% (Fig \ref{property_price_DetachedExisting_g20_obs} and \ref{property_price_DetachedExisting_g20_pred}).

%% file: tables/table_model_test_set_metrics_simulated.tex
%

\begin{sidewaystable*}
\caption{Results on simulated data: summary of test-set mean squared error and maximum monotonicity violation for each model.}\label{table_model_test_set_metrics_simulated}
\begin{tabular*}{\textheight}{@{\extracolsep\fill}lccccccccc}
\toprule%
 & & & & \multicolumn{3}{@{}c@{}}{Mean squared error} & \multicolumn{3}{@{}c@{}}{\makecell{Maximum \\ monotonicity \\ violation}} \\\cmidrule{5-7}\cmidrule{8-10}%
Simulated dataset & Loss$^{1}$ & Model Type & N seeds & Mean & Median & Min & Mean & Median & Max \\
\midrule
Sim-Main & Standard & CoxPH & 1 & 0.065 & 0.065 & 0.065 & \textbf{0} & \textbf{0} & \textbf{0} \\
Sim-Main & LossDyDg & DeepHit & 5 & 0.055 & 0.055 & 0.055 & 0.019 & 0.019 & 0.02 \\
Sim-Main & SUMO & DeepHit & 5 & 0.165 & 0.165 & 0.164 & 0.00039 & 0.00042 & 0.00043 \\
Sim-Main & Standard & GBSA & 5 & 0.101 & 0.101 & 0.101 & 0.12 & 0.12 & 0.12 \\
Sim-Main & Standard & RSF & 5 & 0.086 & 0.086 & 0.086 & 0.16 & 0.16 & 0.16 \\
Sim-Main & LossDyDg & SurvSurf & 5 & \textbf{0.054} & \textbf{0.054} & \textbf{0.053} & \textbf{0} & \textbf{0} & \textbf{0} \\
\midrule
Sim-LackProg & Standard & CoxPH & 1 & \textbf{0.046} & \textbf{0.046} & \textbf{0.046} & \textbf{0} & \textbf{0} & \textbf{0} \\
Sim-LackProg & LossDyDg & DeepHit & 5 & 0.051 & 0.051 & 0.051 & \textbf{0} & \textbf{0} & \textbf{0} \\
Sim-LackProg & SUMO & DeepHit & 5 & 0.076 & 0.076 & 0.075 & 0.0016 & \textbf{0} & 0.0081 \\
Sim-LackProg & Standard & GBSA & 5 & 0.066 & 0.066 & 0.066 & 0.014 & 0.014 & 0.014 \\
Sim-LackProg & Standard & RSF & 5 & 0.058 & 0.058 & 0.058 & 0.06 & 0.06 & 0.07 \\
Sim-LackProg & LossDyDg & SurvSurf & 5 & 0.052 & 0.048 & 0.047 & \textbf{0} & \textbf{0} & \textbf{0} \\
\botrule
\end{tabular*}
\begin{tablenotes}
\item[$^{1}$] The value `Standard' in the Loss column refers to the use of default optimization targets for the models. CoxPH uses negative partial log-likelihood. The other methods have more complex optimization, details can be found through \verb+sksurv+ \citep{RN31}.
\end{tablenotes}
\end{sidewaystable*}

%% file: tables/table_model_test_set_metrics_real.tex
%

\begin{sidewaystable*}
\caption{Results on real-world data: summary of test-set mean $IBI^{ipcw}_{iti}$ over events and maximum monotonicity violation for each model.}\label{table_model_test_set_metrics_real}
\begin{tabular*}{\textheight}{@{\extracolsep\fill}lccccccccc}
\toprule%
 & & & & \multicolumn{3}{@{}c@{}}{\makecell{Mean $IBI^{ipcw}_{iti}$ \\ over events}} & \multicolumn{3}{@{}c@{}}{\makecell{Maximum \\ monotonicity \\ violation}} \\\cmidrule{5-7}\cmidrule{8-10}%
\makecell{Real-world \\ dataset} & Loss$^{1}$ & Model Type & N seeds & Mean & Median & Min & Mean & Median & Max \\
\midrule
RW-TRAE & Standard & CoxPH & 1 & 0.186 & 0.186 & 0.186 & \textbf{0} & \textbf{0} & \textbf{0} \\
RW-TRAE & LossDyDg & DeepHit & 5 & 0.176 & 0.175 & 0.175 & 0.015 & 0.014 & 0.021 \\
RW-TRAE & SUMO & DeepHit & 5 & 0.301 & 0.301 & 0.3 & 0.0033 & 0.0033 & 0.0034 \\
RW-TRAE & Standard & GBSA & 5 & \textbf{0.15} & \textbf{0.15} & \textbf{0.15} & 0.06 & 0.06 & 0.06 \\
RW-TRAE & Standard & RSF & 5 & 0.191 & 0.191 & 0.19 & 0.03 & 0.034 & 0.038 \\
RW-TRAE & LossDyDg & SurvSurf & 5 & 0.155 & 0.155 & 0.155 & \textbf{0} & \textbf{0} & \textbf{0} \\
\midrule
RW-Property & Standard & CoxPH & 1 & 0.085 & 0.085 & 0.085 & \textbf{0} & \textbf{0} & \textbf{0} \\
RW-Property & LossDyDg & DeepHit & 5 & 0.211 & 0.216 & 0.196 & \textbf{0} & \textbf{0} & \textbf{0} \\
RW-Property & SUMO & DeepHit & 5 & 0.293 & 0.293 & 0.293 & 3.7e-06 & 3.7e-06 & 4.1e-06 \\
RW-Property & Standard & GBSA & 5 & 0.12 & 0.12 & 0.12 & 0.073 & 0.073 & 0.073 \\
RW-Property & Standard & RSF & 5 & 0.112 & 0.112 & 0.111 & 0.069 & 0.069 & 0.074 \\
RW-Property & LossDyDg & SurvSurf & 5 & \textbf{0.069} & \textbf{0.07} & \textbf{0.064} & \textbf{0} & \textbf{0} & \textbf{0} \\
\botrule
\end{tabular*}
\begin{tablenotes}
\item[$^{1}$] The value `Standard' in the Loss column refers to the use of default optimization targets for the models. CoxPH uses negative partial log-likelihood. The other methods have more complex optimization, details can be found through \verb+sksurv+ \citep{RN31}.
\end{tablenotes}
\end{sidewaystable*}

%% file: text/discussion.tex
SurvSurf was consistently observed to be one of the best-performing models for predicting the $CIF$s of sequential events from baseline features (or covariates), and was able to identify and utilize the most influential features (Fig \ref{trial_survsurf_feat_perm_perf_change}). The lack of monotonicity violation confirms the theoretically deduced model behavior with respect to changes in $t$ and $g$ (Appendix \ref{secA1}). 

The existence of monotonicity violation in other nonlinear models reflects a lack of constraint on how the $\hat{CIF}$s change with increasing $g$, which was modeled as an extra covariate to allow for the possibility of discovering the association between the $CIF$s and the $g$. Unsurprisingly, RW-Property, the dataset with the largest training set, resulted in the smallest number of (nonlinear) models with non-negligible monotonicity violation, while the smallest training set (RW-TRAE) had the most models with non-negligible violation even after attempts to reduce overfitting (Fig \ref{performance_violation_scatter__all_main_ds}). This shows the importance of mathematically constraining a model to avoid spurious predictions (e.g. the monotonicity constraints in SurvSurf), especially when there is a limited amount of data as commonly seen in e.g. medical applications.

Our attempt in unifying discrete and (continuous) threshold-based sequential events under the same training protocol revealed an interesting challenge in training the models. For discrete events, it may be possible to specify the complete set of events of interest, e.g. when there are a finite number of disease severity grades bounded by an upper limit. However, scenarios involving recurrent events may not have an upper limit, or the upper limit of interest may change due to a shift in research focus. Moreover, when the sequential events are threshold-based and the trajectories consist of continuous measurements, the set of events of interest can be arbitrarily small or large, and is theoretically infinite in size. 

For models that are not linear or SurvSurf, it is difficult to determine which values of $g$ (beyond what is already recorded in each trajectory) the model should be trained on. For all non-\verb|LossDyDg| models, we took the approach of including next event in the sequence which has not occurred by the end of the monitoring period (i.e. maximum $g$ for the trajectory + $\delta_g$)  as a right-censored observation.

However, such a sparse set of $g$ may result in insufficient regularisation of $\hat{CIF}$s with respect to $g$, especially for the less frequently reached $g$'s. This was explored in an additional experiment where all nonlinear and non-SurvSurf models were trained with a denser set of $g$. The denser set of $g$ resulted in improved test-set $MSE$ and $IBS^{ipcw}_{iti}$ and less monotonicity violation for DeepHit and RSF in two of three datasets tested (Supplemental Material \ref{performance_violation_scatter_more_g__all_main_ds}), although none outperformed SurvSurf. On the other hand, a dense set of $g$ may enrich the apparent training data with too many unreached $g$'s so the (nonlinear and non-SurvSurf) models favour overall negative predictions. This was seen in the RW-TRAE dataset where a denser set of $g$ showed a tendency of yielding more false negative predictions (see Supplemental Material \ref{mean_fract_pos_fn_more_g_vs_less_g_DeepHit_RW-TRAE}, \ref{mean_fract_pos_fn_more_g_vs_less_g_GradientBoostingSurvivalAnalysis_RW-TRAE} and \ref{mean_fract_pos_fn_more_g_vs_less_g_RandomSurvivalForest_RW-TRAE}). 

\verb|LossDyDg| represents another attempt in avoiding an arbitrary set of $g$. For DeepHit, it was a significantly better loss function than SUMO in the setting of sequential events. This could be attributed to the fact that all monitoring time-points, as opposed to only the first hitting times in SUMO, were used in computing the loss, hence imposing more precise and accurate constraints on $\hat{CIF}$s. Further improvement seen in SurvSurf, when compared to DeepHit-LossDyDg, was caused by the enhanced extrapolation ability of SurvSurf as a result of the monotonicity guarantee. Specifically, a local association between $g$ and $\hat{CIF}$ encoded by the $\Delta \hat{CIF}_g$ term in Equation (\ref{DyDg_essense}) implies that (given the same baseline features) the difference in the $\hat{CIF}$s between any smaller $g$'s and any larger $g$'s can not be smaller in magnitude than $\Delta \hat{CIF}_g$ and must be in the same direction. So, the locally defined loss helps inform the global $\hat{CIF}$ through the monotonicity property of SurvSurf, without having to present additional data-points (i.e. a dense event set) to the model.

In addition, SurvSurf, being a neural network, can be flexibly applied on top of other neural network architectures that can process data with complex non-tabular structures, such as imaging, audio, and free text data. This, together with the ability to handle missing intermediate events, makes SurvSurf compatible to many real-world applications.

Finally, this study only used baseline (i.e., not time-dependent) features for prediction. The structure of our model, however, does not prohibit the inclusion of time-dependent features. Extensions of this work may consider incorporation of time-dependent features. Future work may additionally explore the impact of competing events, which was treated as right-censoring and was not explicitly modeled in the current work.

%% file: text/conclusion.tex
We have demonstrated in this work that the proposed SurvSurf model is a superior method for predicting (from baseline predictors) the distribution of event times (specifically first hitting time) for sequential events that are either discrete or defined by continuous thresholds. 
Not only did the model achieve competitive performance on a variety of datasets by incorporating implied facts about omitted intermediate events based on knowledge of sequential events. The model has the unique advantage that it would never violate the monotonic relationship between the $CIF$s of an earlier-order event and a later-order event for sequential events, providing guarantee against spurious predictions. We consider SurvSurf to be a robust and reliable choice for sequential event modeling, offering significant advancement in predictive performance and theoretical soundness.

%% file: text/proof__layer_spec_monotone.tex
\section{Proof of monotonicity guarantee of SurvSurf}\label{secA1}

\label{sec:proof__layer_monotonicity}
This section provides theorems to show that $\frac{\partial M}{\partial t} \geq 0$ and $\frac{\partial M}{\partial g} \leq 0$ for the network output $M(t, g, z_0) = M_{raw}(t, g, z_0) - M_{raw}(t=0, g, z_0)$ where $M_{raw}$ is the output from the last activation function in the network with the $k$ th layer defined according to Section \ref{sec:proposed_model_spec}:
\begin{equation}
\label{eq:z_k_simplified}
z_{k}(t, g, z_{k-1}, z_0) = \sigma_k(\alpha_kt + \gamma_k \diamond h_k(t, g) + A_kz_{k-1} + f_k(z_0) + \beta_k)
\end{equation}
where $z_0$ is a function of $x$, $\diamond$ is a custom operation that is defined as a scalar multiplication if $h_k(t, g)$ returns a scalar and an element-wise product if $h_k(t, g)$ returns a vector. $h_k(t, g)$ is a function that is weakly monotonic (increasing) in $t$ and and (decreasing) in $g$:
\begin{equation}
\begin{split}
h_k(t, g) = [\sigma^{sigm}(
M_k^{time}t+c_k^{time})-\\ \sigma^{sigm}(c_k^{time})]\diamond \sigma^{sigm}({-M}_k^{grade}g+c_k^{grade})
\end{split}
\end{equation}
where $\sigma^{sigm}$ is the sigmoid activation function. Note that $h(t=0,g) = 0$ and $h(t \geq 0, g) \geq 0$.

\begin{theorem}$\frac{\partial M}{\partial t} = \frac{\partial M_{raw}}{\partial t}$ and $\frac{\partial M}{\partial g} = \frac{\partial M_{raw}}{\partial g}$
\label{sec:proof_simplify_gradient}
    \begin{proof} Since $M(t, g, z_0) = M_{raw}(t, g, z_0) - M_{raw}(t=0, g, z_0)$ and $M_{raw}(t=0, g, z_0) = z_{k_{max}}(t=0, g, z_{k_{max} - 1}, z_0)$, and by Equation \eqref{eq:z_k_simplified}
    \begin{equation}
    z_{k}(t=0, g, z_{k-1}, z_0) = \sigma_k(A_kz_{k-1} + f_k(z_0) +\beta_k),
    \end{equation}
    then
    \begin{equation}
    z_{1}(t=0, g, z_0, z_0) = \sigma_{1}(A_1z_{0} + f_1(z_0)+\beta_1),
    \end{equation}
    
    Since $z_{1}(t=0, g, z_0, z_0)$ is constant in $t$ and $g$,  $z_{k_{max}}(t=0, g, z_{k_{max} -1}, z_0)$ will also be constant in $t$ and $g$, by the chain rule and by induction. Thus,  $\frac{\partial M_{raw}(t=0)}{\partial t}=0$ and $\frac{\partial M_{raw}(t=0)}{\partial g}=0$. Therefore,
    \begin{equation}
        \frac{\partial M}{\partial t} = \frac{\partial M_{raw}}{\partial t} - \frac{\partial M_{raw}(t=0)}{\partial t} = \frac{\partial M_{raw}}{\partial t}
    \end{equation}
    \begin{equation}
        \frac{\partial M}{\partial g} = \frac{\partial M_{raw}}{\partial g} - \frac{\partial M_{raw}(t=0)}{\partial g} = \frac{\partial M_{raw}}{\partial g}
    \end{equation}
    \end{proof}
    
\end{theorem}

\begin{theorem}[monotonicity in $t$]$\frac{\partial M_{raw}}{\partial t} \geq 0$
\label{sec:proof_monotone_t}
    \begin{proof}By the chain rule, $\frac{\partial z_{k}}{\partial t} = \frac{\partial \sigma_{k}}{\partial u_k}\cdot\frac{\partial u_k}{\partial t}$ where $u_k = \alpha_kt + \gamma_k \diamond h_k(t, g) + A_kz_{k-1} + f_k(z_0) + \beta_k$,  
    
    \begin{equation}
    \frac{\partial z_{k}}{\partial t} =  \frac{\partial \sigma_{k}}{\partial u_k}\cdot(\alpha_k + \gamma_k\diamond\frac{\partial h_k(t,g)}{\partial t} + A_k\frac{\partial z_{k-1}}{\partial t})
    \end{equation}
    
    Given that 
    \begin{equation}
    z_{1} = \sigma_{1}(\alpha_1t + \gamma_1\diamond h_1(t, g) + A_1z_0 + f_1(z_0) + \beta_1)
    \end{equation}
    then 
    \begin{equation}
        \frac{\partial z_{1}}{\partial t} = \frac{\partial \sigma_{1}}{\partial u_1}\cdot(\alpha_1 + \gamma_1\diamond\frac{\partial h_1(t,g)}{\partial t})
    \end{equation}
    Since $\frac{\partial \sigma_{1}}{\partial u_1} > 0$ ($\sigma_{1} = tanh$ or the identity function), $g > 0$, $\frac{\partial h_k(t,g)}{\partial t} \geq 0$, and  $\alpha_{1}$, $\gamma_{1}$ are both point-wise non-negative, it can be concluded that $\frac{\partial z_{1}}{\partial t} \geq 0$. Similarly, as $A_k$ is point-wise non-negative (so are $\alpha_k$ and $\gamma_k$), $\frac{\partial z_{k+1}}{\partial t} \geq 0$ when $k = 1$ and hence for all $k > 1$ as well. Therefore $\frac{\partial M_{raw}}{\partial t} \geq 0$.
    \end{proof}
\end{theorem}

\begin{theorem}[monotonicity in $g$]$\frac{\partial M_{raw}}{\partial g} \leq 0$\label{sec:proof_monotone_g}
    \begin{proof} This proof proceeds in a similar way to that for Theorem
    \ref{sec:proof_monotone_t}.
    
    By the chain rule, $\frac{\partial z_{k}}{\partial g} = \frac{\partial \sigma_{k}}{\partial u_k}\cdot \frac{\partial u_k}{\partial g}$ where $u_k = \alpha_kt + \gamma_k\diamond h_k(t, g) + A_kz_{k-1} + f_k(z_0) + \beta_k$.  
    
    \begin{equation}
    \frac{\partial z_{k}}{\partial g} =  \frac{\partial \sigma_{k}}{\partial u_k}\cdot(\gamma_k\diamond \frac{\partial h_k(t,g)}{\partial g} + A_k\frac{\partial z_{k-1}}{\partial g})
    \end{equation}

    Given that \begin{equation}
    z_{1} = \sigma_{1}(\alpha_{1}t + \gamma_{1}\diamond h_1(t, g) + f_1(z_0) + \beta_{1})
    \end{equation},
    then 
    \begin{equation}
    \frac{\partial z_{1}}{\partial g} =  \frac{\partial \sigma_{1}}{\partial u_1}\cdot(\gamma_k\diamond \frac{\partial h_1(t,g)}{\partial g})
    \end{equation}
    Since $\frac{\partial \sigma_{1}}{\partial u_1} > 0$ ($\sigma_{1} = tanh$ or the identity function), $g > 0$, $t \geq 0$, $\frac{\partial h_k(t,g)}{\partial g} \leq 0$, and $\gamma_{1}$ is point-wise non-negative, it can be concluded that $\frac{\partial z_{1}}{\partial g} \leq 0$. Similarly, as $A_k$ is point-wise non-negative (so is $\gamma_k$), $\frac{\partial z_{k+1}}{\partial g} \leq 0$ when $k = 1$ and hence for all $k > 1$ as well. Therefore $\frac{\partial M_{raw}}{\partial g} \leq 0$.
    \end{proof}
\end{theorem}



\begin{theorem}
    For the SurvSurf model, the network output is monotonically increasing in $t$ and decreasing in $g$, i.e. $\frac{\partial M}{\partial t} \geq 0$ and $\frac{\partial M}{\partial g} \leq 0$.
    \begin{proof} By Theorem \ref{sec:proof_simplify_gradient}, Theorem \ref{sec:proof_monotone_t} and Theorem \ref{sec:proof_monotone_g}, $\frac{\partial M_{raw}}{\partial t} \geq 0$ and $\frac{\partial M_{raw}}{\partial g} \leq 0$. Therefore by \hyperref[sec:lemma_M1]{Lemma M1}, the result follows.
    \end{proof}
\end{theorem}

%% file: text/supp.tex
\setcounter{figure}{0}
\renewcommand{\thefigure}{S\arabic{figure}}

\begin{figure}[!ht]
    \centering
    \includegraphics[width=0.9\linewidth]{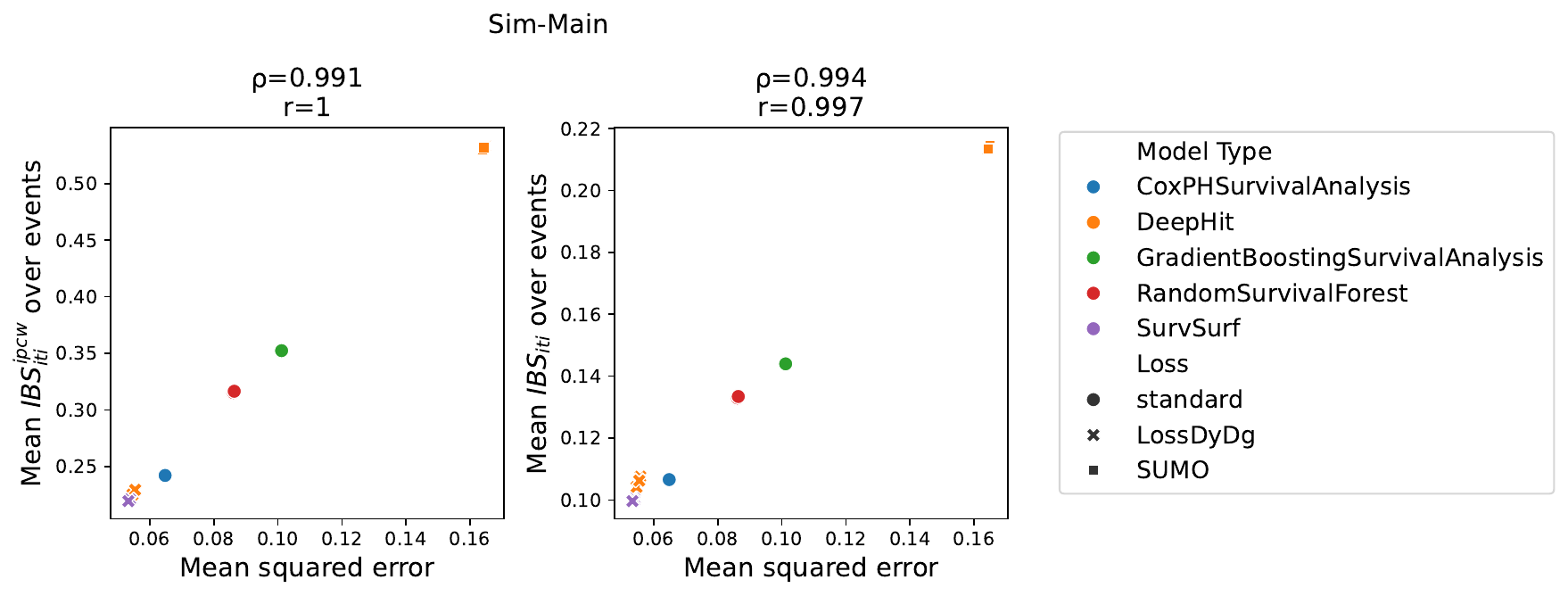}
    \caption{Test-set result: Spearman's rank correlation ($\rho$) and Pearson's R (r) between mean squared error,  $IBS^{ipcw}_{iti}$ (left) and $IBS_{iti}$ (right) on Sim-Main dataset.}\label{MSE_vs_ipcwless_IBS_iti_main_sim}
\end{figure}

\begin{figure}[ht]
    \centering
    \includegraphics[width=0.9\linewidth]{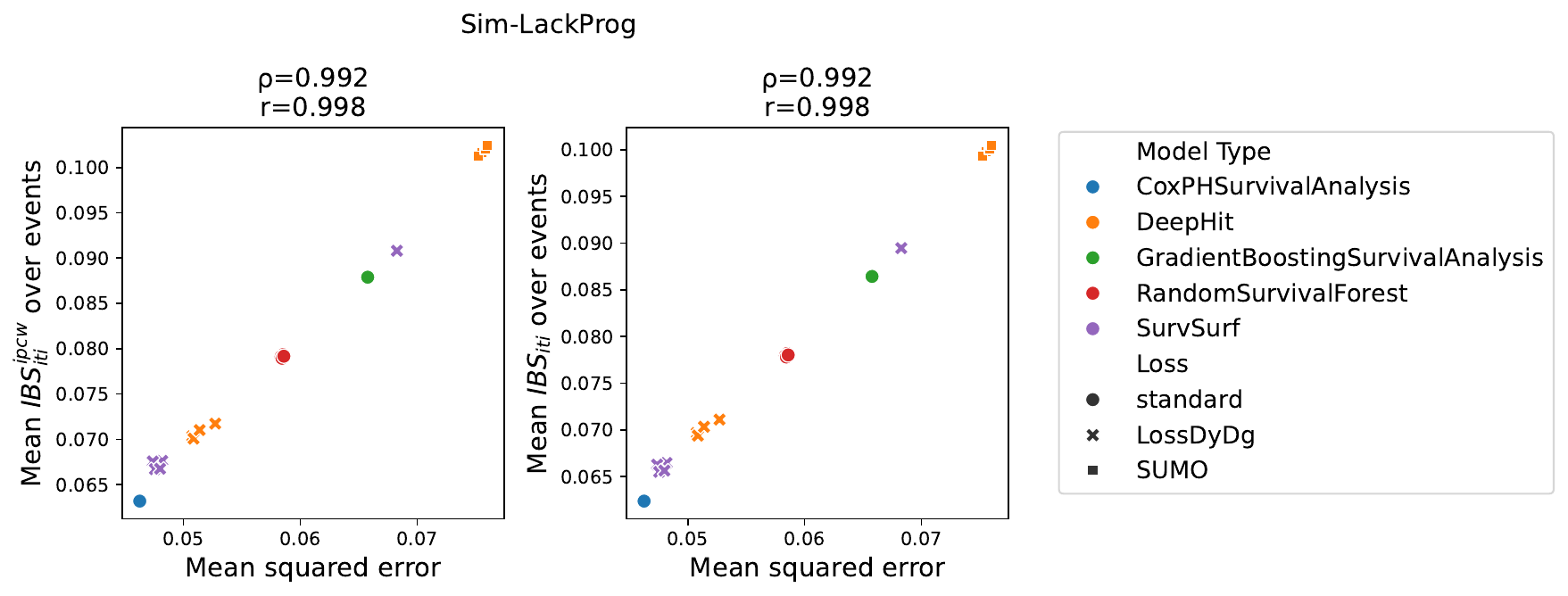}
    \caption{Test-set result: Spearman's rank correlation ($\rho$) and Pearson's R (r) between mean squared error,  $IBS^{ipcw}_{iti}$ (left) and $IBS_{iti}$ (right) on Sim-LackProg dataset.}\label{MSE_vs_ipcwless_IBS_iti_less_prog_sim}
\end{figure}

\begin{figure}
    \centering
    \includegraphics[width=1\linewidth]{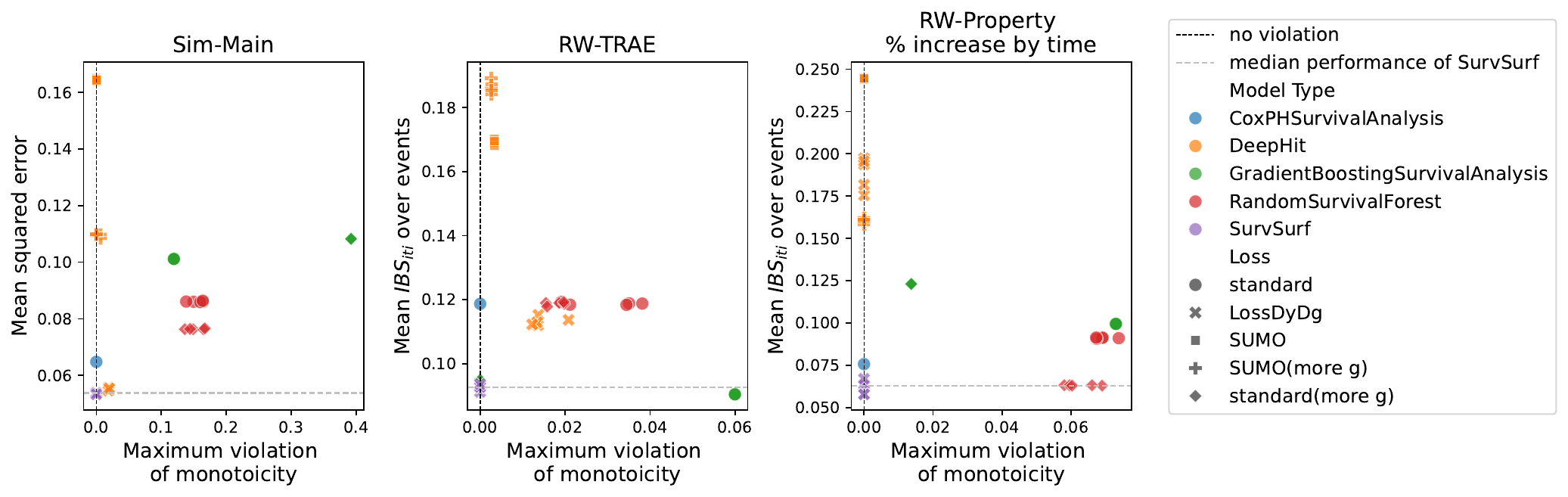}
    \caption{Model performance (mean squared error and $IBS_{iti}$) and the extent of monotonicity violation for the SurvSurf and the benchmarking models (including nonlinear non-SurvSurf models trained with a denser set of $g$) in the test set of different datasets. Points with the same color and marker are the same model trained by initializing with different random seeds. Points closer to the bottom-left corner represent models with better performance and less monotonicity violation. The denser set of $g$ consists of the observed $g$s in each trajectory and equally spaced $g$s between the maximum observed $g$ in each trajectory and the upper bound used in computing model evaluation metrics. The spacing in $g$ is the same as that for model evaluation. }\label{performance_violation_scatter_ipcwless__all_main_ds}
\end{figure}

\begin{figure}
    \centering
    \includegraphics[width=1\linewidth]{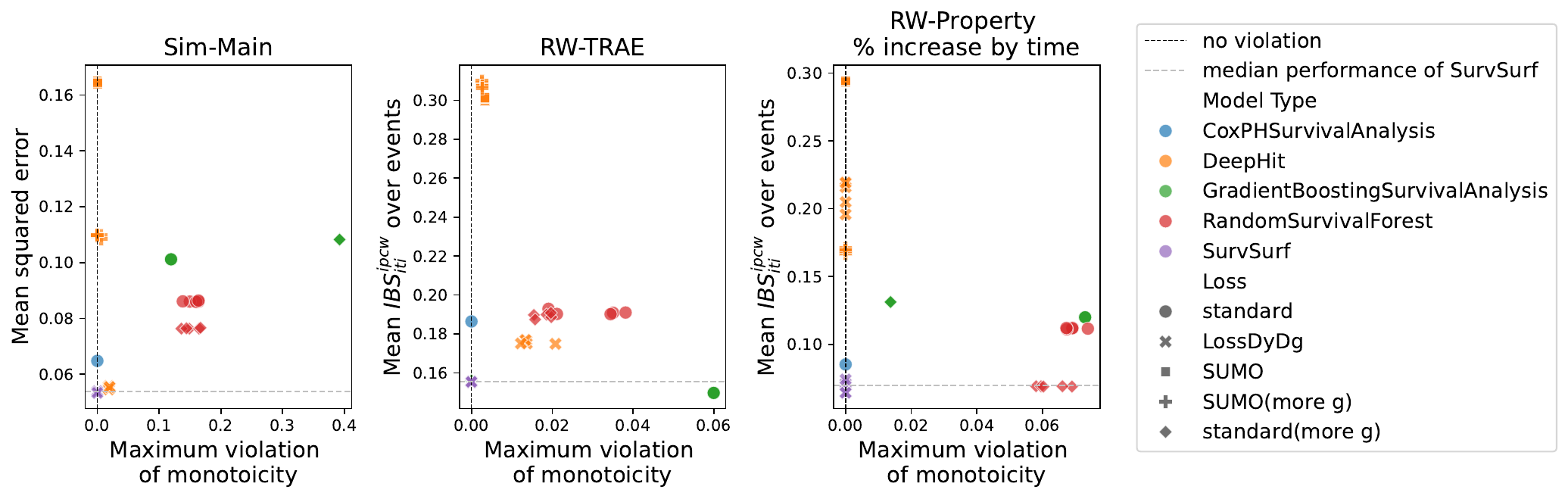}
    \caption{Model performance (mean squared error and $IBS^{ipcw}_{iti}$) and the extent of monotonicity violation for the SurvSurf and the benchmarking models (including nonlinear non-SurvSurf models trained with a denser set of $g$) in the test set of different datasets. Points with the same color and marker are the same model trained by initializing with different random seeds. Points closer to the bottom-left corner represent models with better performance and less monotonicity violation. The denser set of $g$ consists of the observed $g$s in each trajectory and equally spaced $g$s between the maximum observed $g$ in each trajectory and the upper bound used in computing model evaluation metrics. The spacing in $g$ is the same as that for model evaluation.}\label{performance_violation_scatter_more_g__all_main_ds}
\end{figure}  

\begin{figure}
    \centering
    \includegraphics[width=1\linewidth]{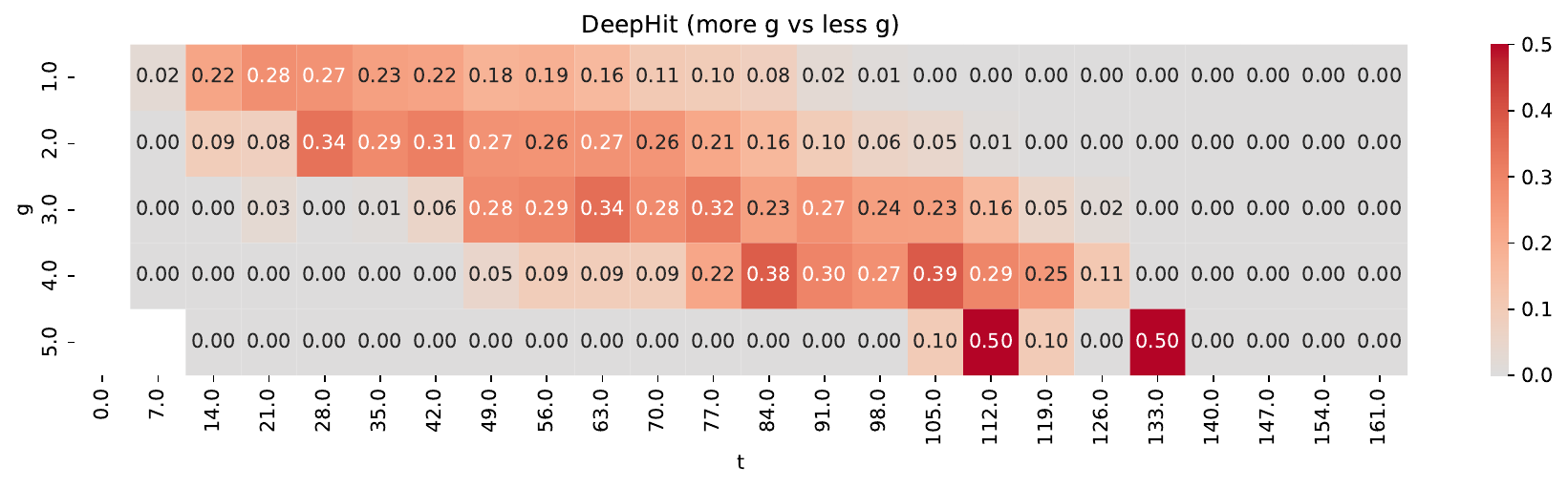}
    \caption{Change in the number of false negative predictions (with probability threshold 0.5) as a fraction of all positive observations after using a denser set of $g$ for model training. Each number is an average across 5 DeepHit models each trained with a different initialization seed.}\label{mean_fract_pos_fn_more_g_vs_less_g_DeepHit_RW-TRAE}
\end{figure}  

\begin{figure}
    \centering
    \includegraphics[width=1\linewidth]{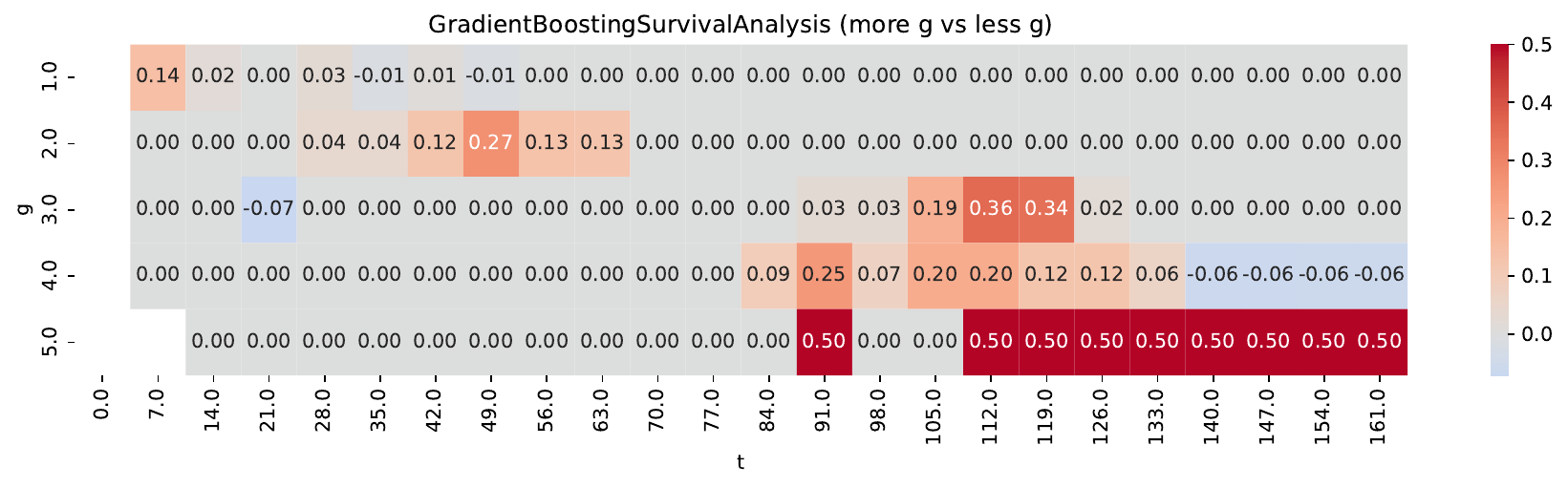}
    \cprotect\caption{Change in the number of false negative predictions (with probability threshold 0.5) as a fraction of all positive observations after using a denser set of $g$ for model training. Each number is an average across 5 \verb+GradientBoostingSurvivalAnlaysis+ models each trained with a different initialization seed on the RW-TRAE dataset.}\label{mean_fract_pos_fn_more_g_vs_less_g_GradientBoostingSurvivalAnalysis_RW-TRAE}
\end{figure}  

\begin{figure}
    \centering
    \includegraphics[width=1\linewidth]{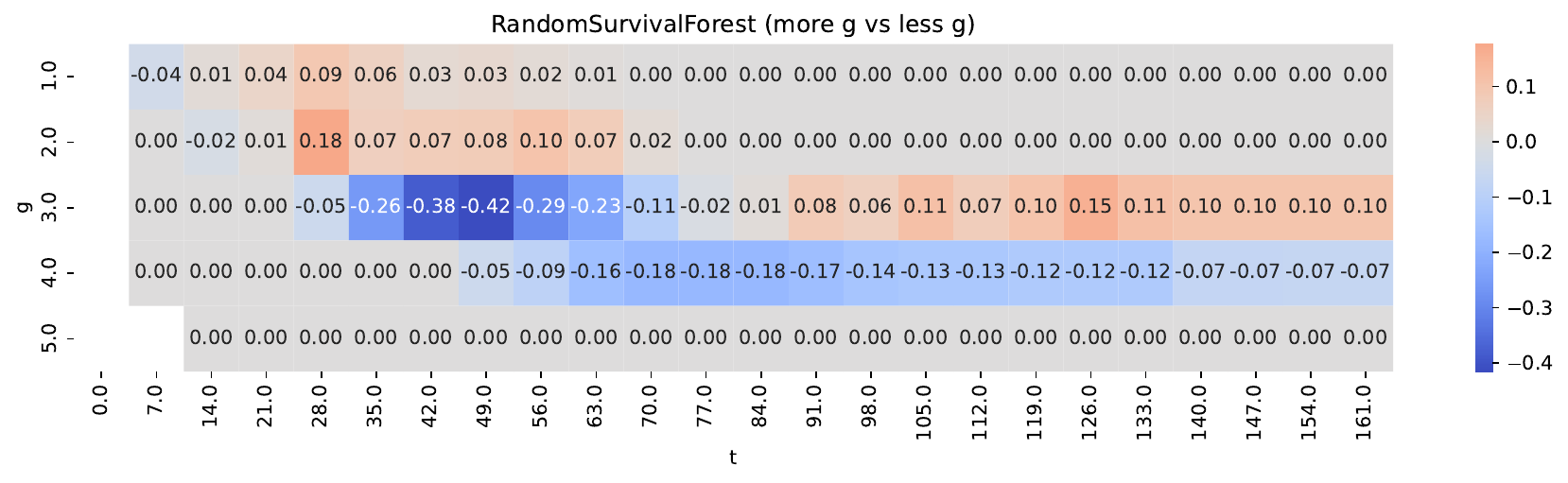}
    \cprotect\caption{Change in the number of false negative predictions (with probability threshold 0.5) as a fraction of all positive observations after using a denser set of $g$ for model training. Each number is an average across 5 \verb+RandomSurvivalForest+ models each trained with a different initialization seed on the RW-TRAE dataset.}\label{mean_fract_pos_fn_more_g_vs_less_g_RandomSurvivalForest_RW-TRAE}
\end{figure}  